\pdfoutput=1

\documentclass[11pt]{article}

\usepackage[preprint]{acl}

\usepackage{times}
\usepackage{latexsym}

\usepackage[T1]{fontenc}

\usepackage[utf8]{inputenc}

\usepackage{microtype}

\usepackage{inconsolata}

\usepackage{graphicx}

\usepackage[utf8]{inputenc} 
\usepackage[T1]{fontenc}    
\usepackage{hyperref}       
\usepackage{url}            
\usepackage{booktabs}       
\usepackage{amsfonts}       
\usepackage{nicefrac}       
\usepackage{microtype}      
\usepackage{xcolor}         
\usepackage{amssymb}
\usepackage{bm}
\usepackage{amsmath}
\usepackage{comment}
\usepackage{tikz}
\usepackage{algorithm}
\usepackage{algpseudocode}
\usepackage{caption}
\usepackage{subcaption}
\usepackage{float}
\usepackage{pifont}
\usepackage{ulem}

\title{Geometric Interpretation of Layer Normalization and \\a Comparative Analysis with RMSNorm}

\author{Akshat Gupta, Atahan Ozdemir, Gopala Anumanchipalli \\
        UC Berkeley \\
        \texttt{\{akshat.gupta, atahanozdemir, gopala\}@berkeley.edu}}

\begin{document}
\maketitle
\begin{abstract}


This paper presents a novel geometric interpretation of LayerNorm and explores how LayerNorm influences the norm and orientation of hidden vectors in the representation space. With these geometric insights, we prepare the foundation for comparing LayerNorm with RMSNorm. We show that the definition of LayerNorm is innately linked to the uniform vector, defined as $\boldsymbol{1} = [1, 1, 1, 1, \cdots, 1]^T \in \mathbb{R}^d$. We then show that the standardization step in LayerNorm can be understood in three simple steps: (i) remove the component of a vector along the uniform vector, (ii) normalize the remaining vector, and (iii) scale the resultant vector by $\sqrt{d}$, where $d$ is the dimensionality of the representation space. We also provide additional insights into how LayerNorm operates at inference time. Finally, we compare the hidden representations of LayerNorm-based LLMs with models trained using RMSNorm and show that all LLMs naturally operate orthogonal to the uniform vector at inference time, that is, on average they do not have a component along the uniform vector during inference. This presents the first mechanistic evidence that removing the component along the uniform vector in LayerNorm is a redundant step. These results advocate for using RMSNorm over LayerNorm which is also more computationally efficient.

\end{abstract}

\setlength{\belowcaptionskip}{-15pt}
\begin{figure}[t]
        \centering
        \includegraphics[width=\columnwidth]{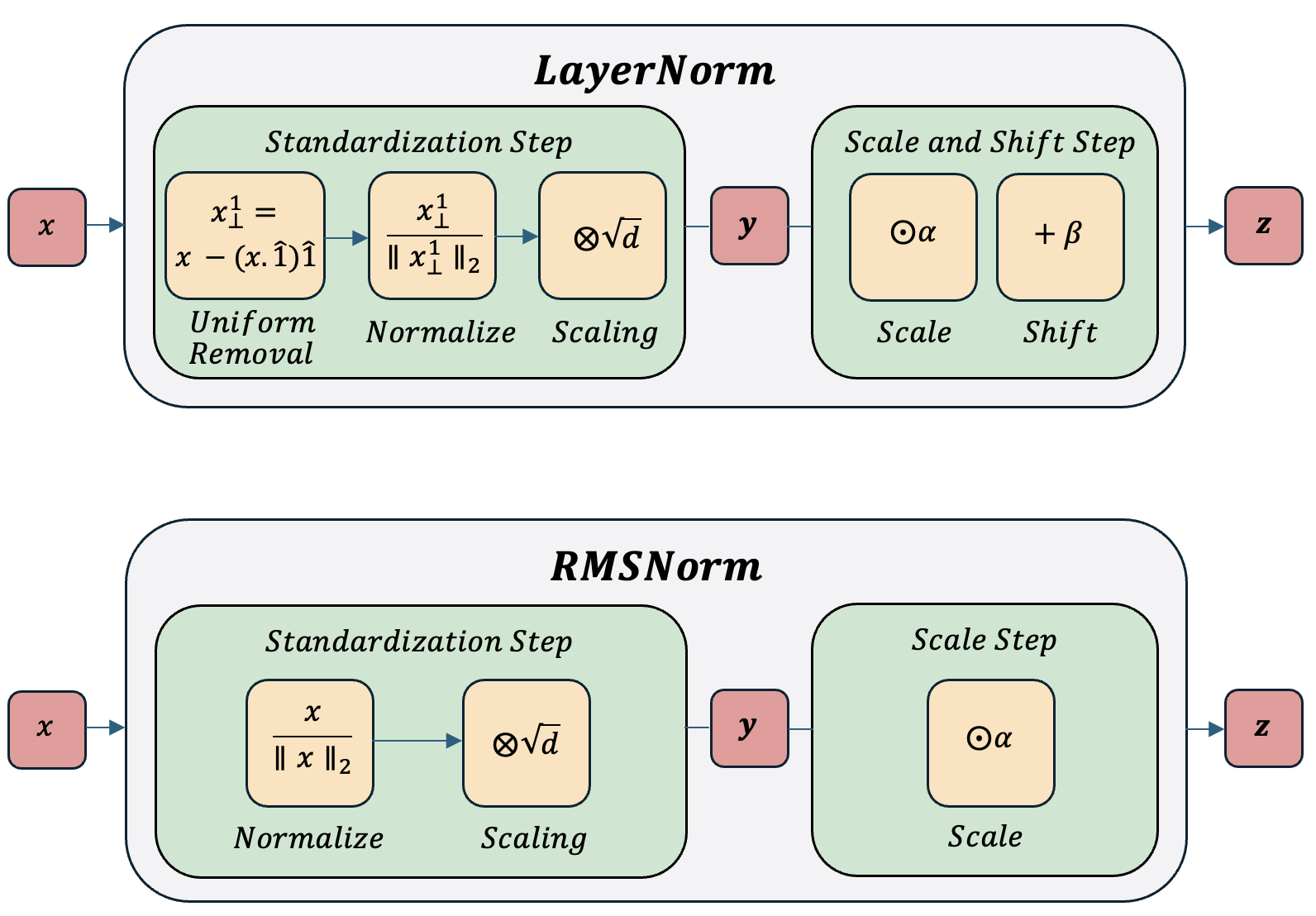}
    \caption{A diagrammatic explanation of LayerNorm and RMSNorm.}
    \label{fig:lndiagram}
\end{figure}

\section{Introduction}



The transformer architecture \citep{transformer} has been the cornerstone of most recent advances in artificial intelligence and has rapidly become the architecture of choice in natural language processing, computer vision, speech, and various other domains. Layer normalization, a crucial yet often overlooked component of the transformer architecture, plays an integral role in stabilizing the training process in transformer-based models. Introduced by \citet{layernorm}, layer normalization standardizes the features at each layer, adjusting and scaling the activations to have zero mean and unit variance within a vector. This normalization is performed independently for each hidden vector in contrast to batch normalization \citep{batchnorm}, which relies on statistics (mean and variance) from a batch of data points. Layer normalization is especially effective when working with long sequences of variable lengths. While the benefits of layer normalization, such as improved training speed and better convergence, are well documented \citep{layernorm, understanding-layernorm, rmsnorm, prermsnorm}, its specific effects on the hidden vectors within a model and the global properties of the resulting representations remain surprisingly underexplored.

\begin{figure*}[t]
\begin{center}
    \begin{subfigure}[b]{0.32\textwidth}
        \centering
        \includegraphics[width=\textwidth]{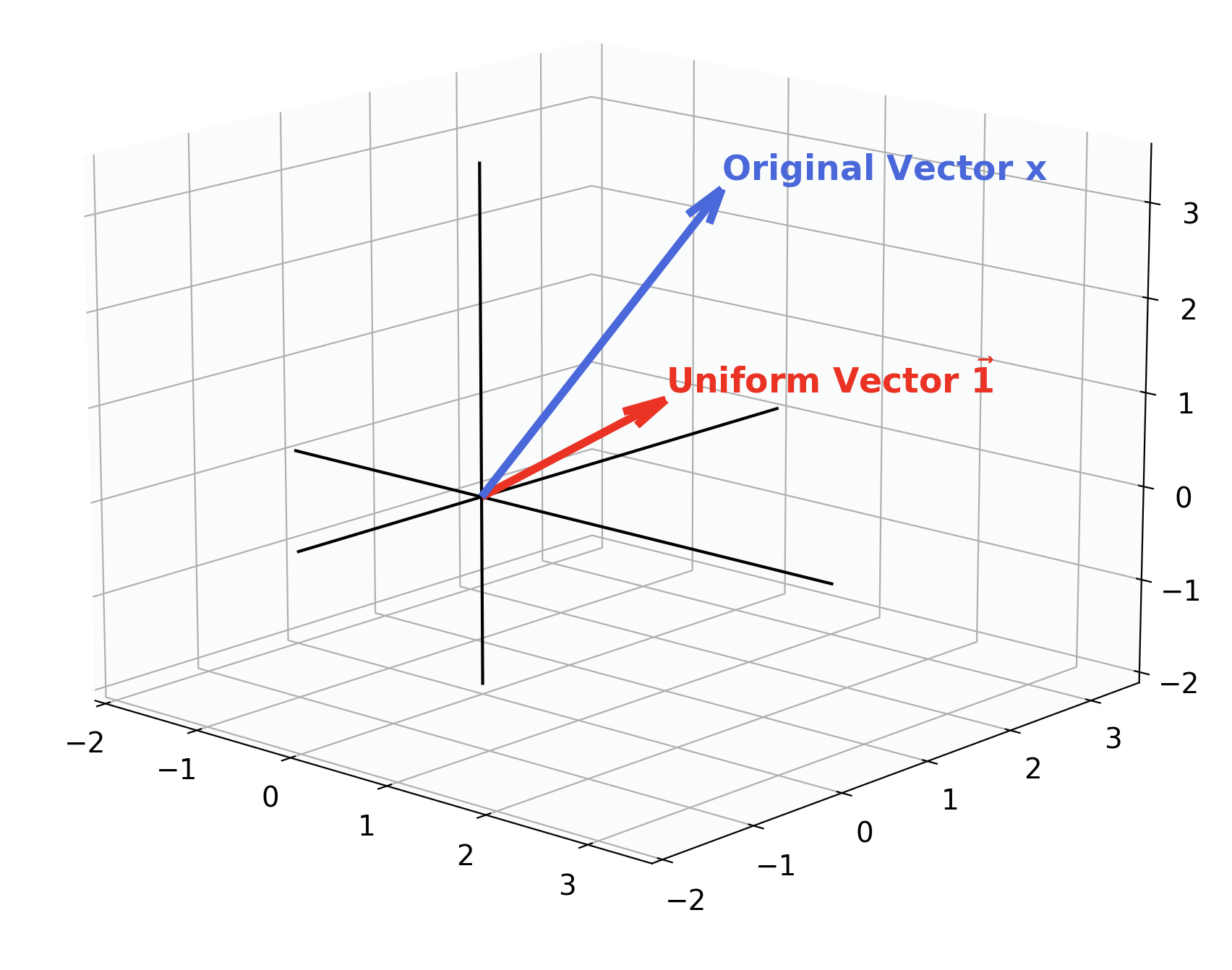}
        \caption{This figure shows the original vector (in blue) and the uniform vector (in red) in a 3-D space.\\}\label{fig:lnprocess_a}
    \end{subfigure}
    \hfill
    \vspace{0.5cm}
    \begin{subfigure}[b]{0.32\textwidth}
        \centering
        \includegraphics[width=\textwidth]{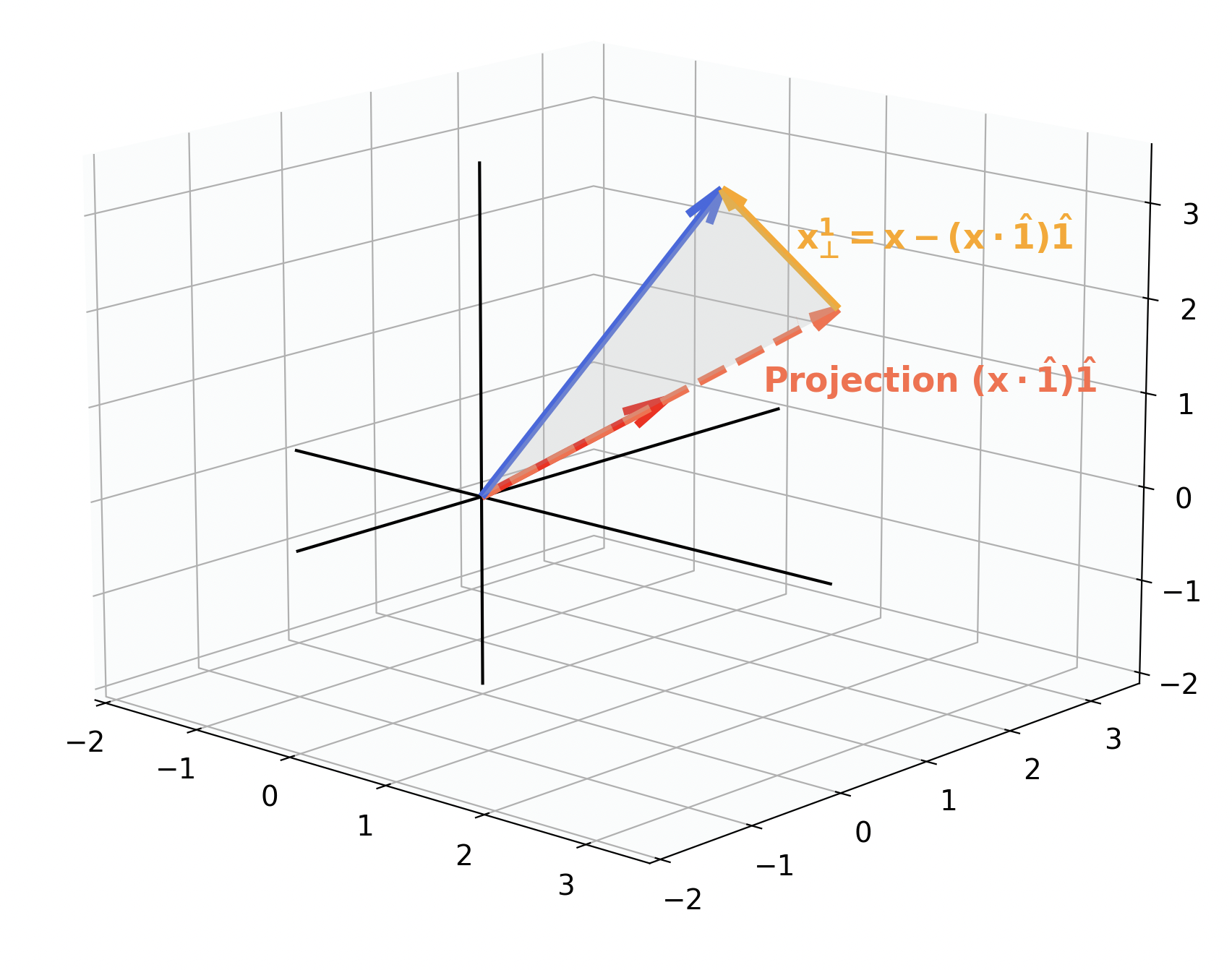}
        \caption{This figure shows the projection of the original vector along the uniform vector, $(\bm{x}.\hat{\bm{1}}) \hat{\bm{1}}$, and the remaining component, $\bm{x^{\boldsymbol{1}}_{\perp}}$.}\label{fig:lnprocess_b}
    \end{subfigure}
    \hfill
    \begin{subfigure}[b]{0.32\textwidth}
        \centering
        \includegraphics[width=\textwidth]{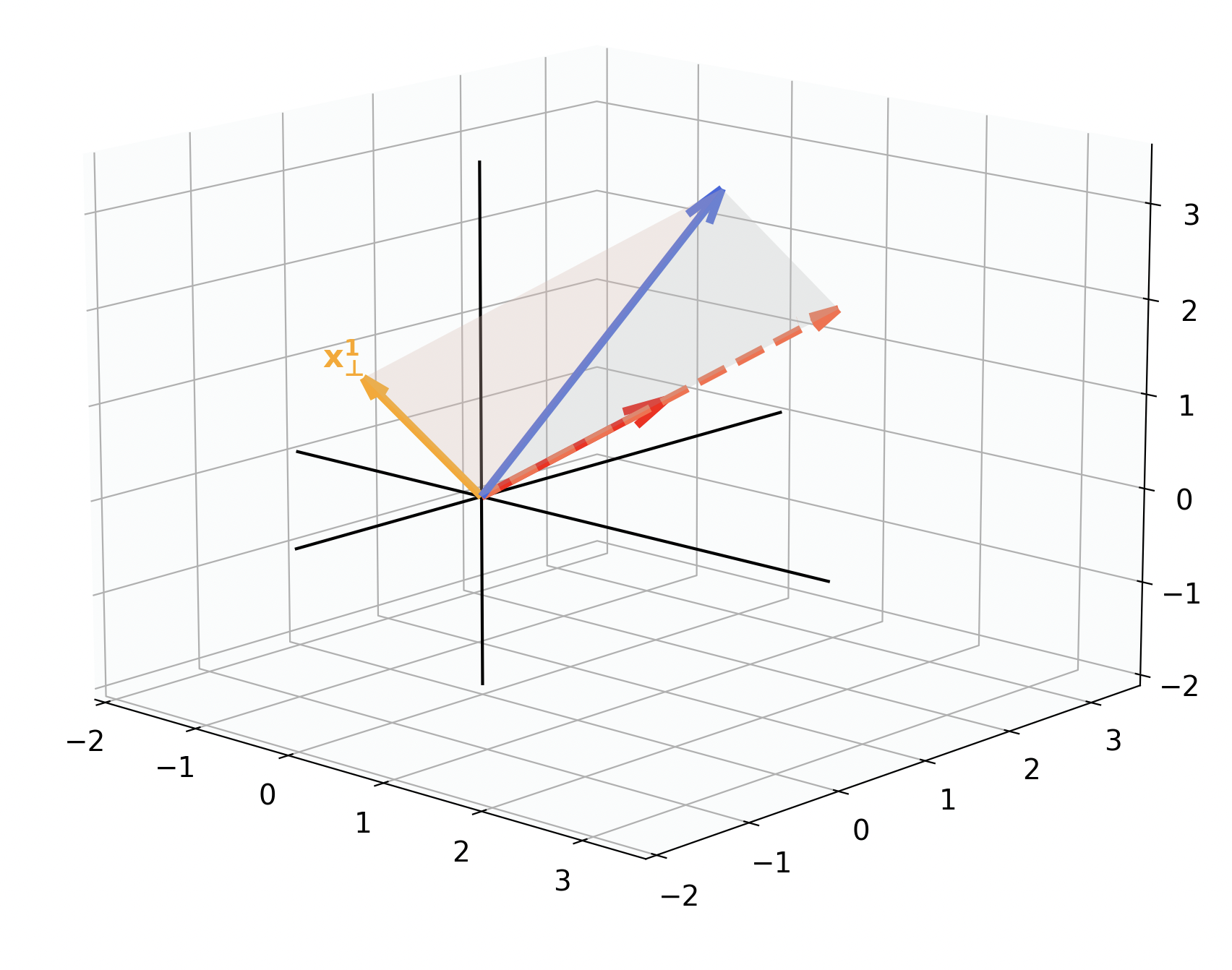}
        \caption{The component of the original vector, after removing the projection along the uniform vector, is kept ( $\bm{x^{\boldsymbol{1}}_{\perp}}$, shown in yellow). }\label{fig:lnprocess_c}
    \end{subfigure}
    \hfill
    \begin{subfigure}[b]{0.32\textwidth}
        \centering
        \includegraphics[width=\textwidth]{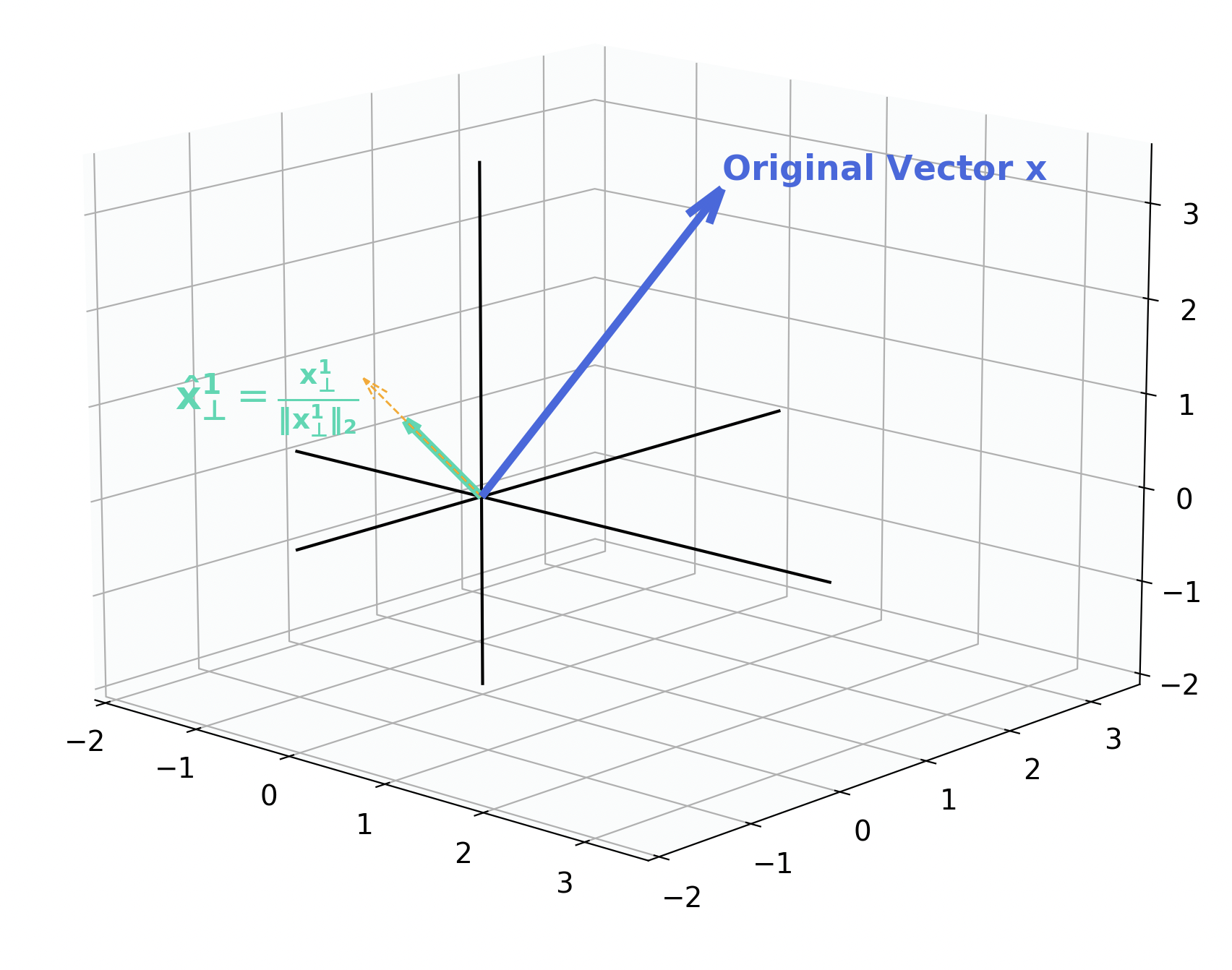}
        \caption{Normalizing perpendicular component to the uniform vector ($\bm{x^{\boldsymbol{1}}_{\perp}}$) to unit norm.\\}\label{fig:lnprocess_d}
    \end{subfigure}
    \hfill
    \begin{subfigure}[b]{0.32\textwidth}
        \centering
        \includegraphics[width=\textwidth]{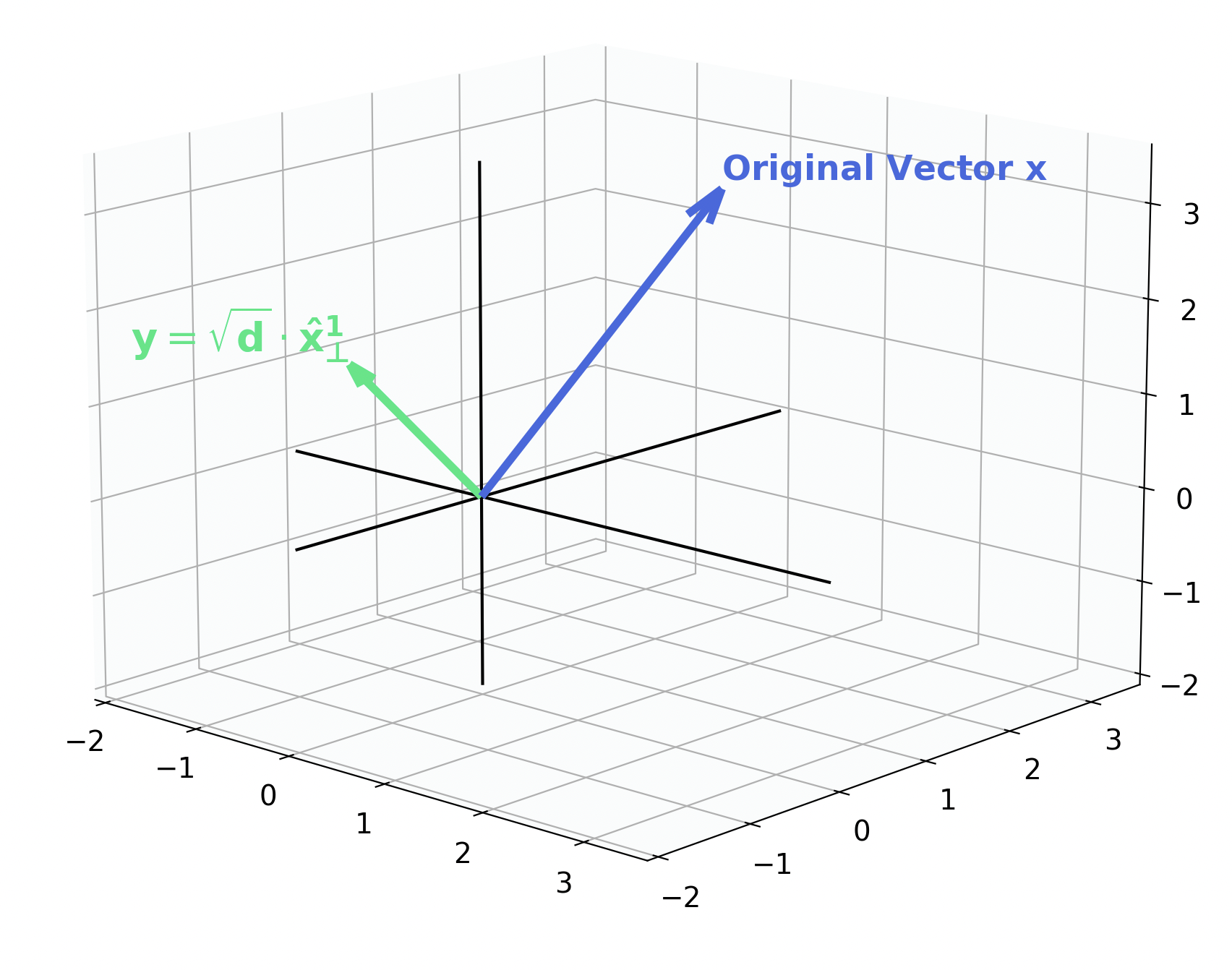}
        \caption{Scale of resultant vector by $\sqrt{d}$, where $d$ is the dimensionality of the representation space.\\}\label{fig:lnprocess_e}
    \end{subfigure}
    \hfill
    \begin{subfigure}[b]{0.32\textwidth}
        \centering
        \includegraphics[width=\textwidth]{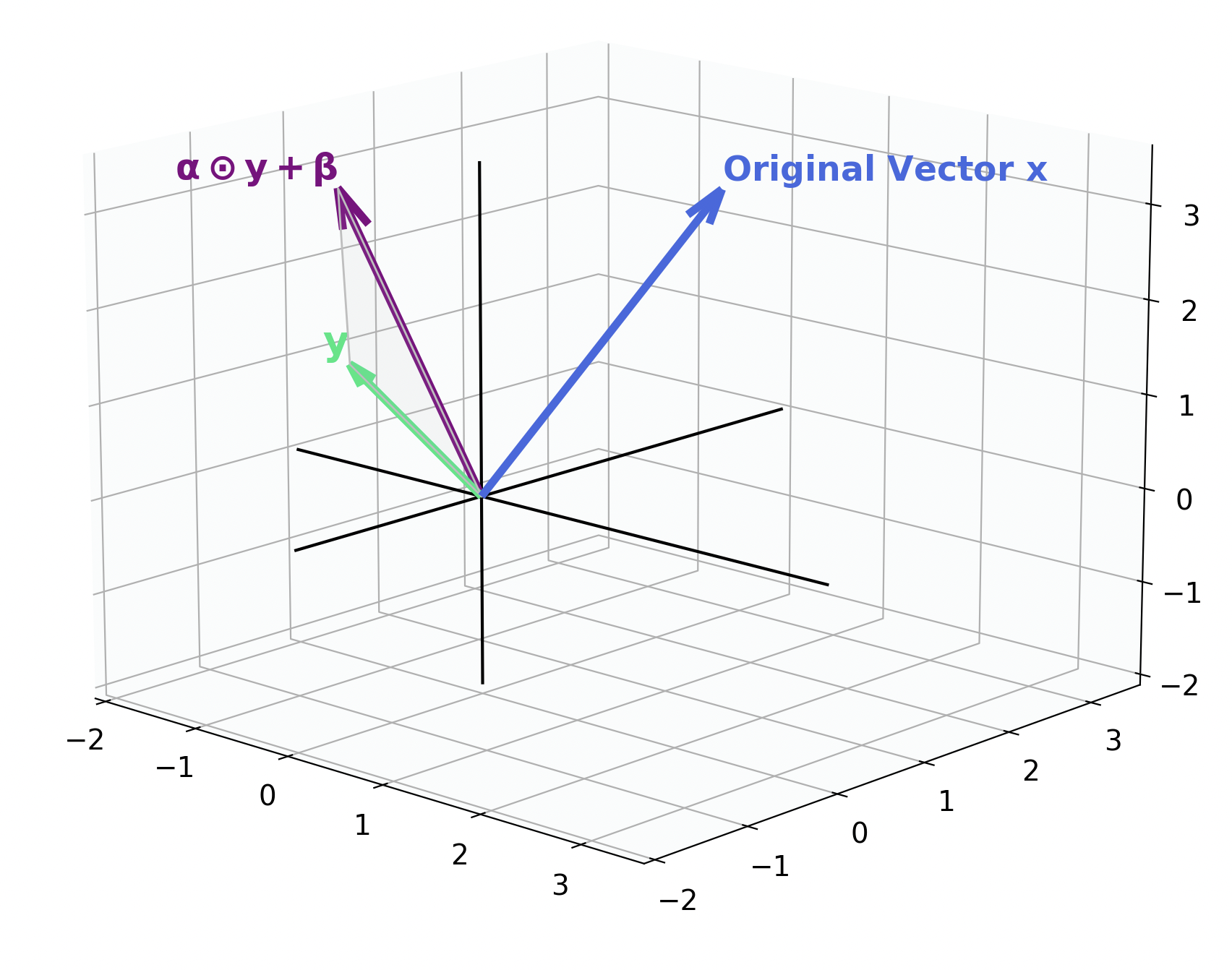}
        \caption{Finally, the scale-and-shift step, which scales and shifts the resulting vector according to learnt parameters.\\}
    \end{subfigure}\label{fig:lnprocess_f}
\end{center}
\caption{Visualization of LayerNorm operation on a random original vector}
\label{fig:lnprocess}
\end{figure*}

In this paper, we first discuss the global effects of layer normalization on a vector. The conventional explanation of the LayerNorm operation is usually as follows: \textit{standardize each vector by subtracting the mean of its elements, divide by the standard deviation}. While this is an accurate procedural definition, we ask a more global question: \textbf{How does the LayerNorm operation transform a vector in representation space?} 

We present a novel interpretation of LayerNorm and show that it can be understood in three steps: (i) throw away the component of the given vector along $\boldsymbol{1} = [1, 1, 1, 1, \cdots, 1]^T \in \mathbb{R}^d$, (ii) normalize the resultant vector; and (iii) scale the resulting vector by $\sqrt{d}$, where $d$ is the dimensionality of the vector space. This process is illustrated in Figure \ref{fig:lnprocess}. This operation throws away the information along $\boldsymbol{1} = [1, 1, 1, 1, \cdots 1]^T$, which we call the uniform vector, indicating that the information along the uniform vector may not be important.

We then present the property of ``irreversibility," where we show that the information lost in the LayerNorm process cannot be recovered. These results naturally lead to discussion about the importance of information along the uniform vector, which is removed by LayerNorm irreversibly, and its comparison with RMSNorm \citep{rmsnorm}, a variant of LayerNorm that doesn't remove the component along the uniform vector and is used to train the latest Llama models \citep{llama, llama2, llama3}. Figure \ref{fig:lndiagram} shows the difference between the two normalization methods. To empirically justify removing the components of hidden vectors along the uniform vectors, the hidden representations should have non-trivial components along the uniform vector. 

We empirically show that both LayerNorm-based models naturally produce hidden representations orthogonal to the uniform vector during inference time, thus rendering the removal of the uniform vector in LayerNorm inconsequential in practice. This is true even for RMSNorm-based models, where hidden vectors on average operate othogonal to the uniform vector. With these results, we provide both theoretical and empirical justifications for the redundancy of removing the component of hidden representations along the uniform vector, thus supporting the usage of RMSNorm over LayerNorm, which is also more computationally efficient.

To summarize, in this paper, we make the following contributions:

\vspace{-5pt}

\begin{enumerate}
\item Provide a simple geometric interpretation of the layer normalization process that offers an intuitive explanation of how LayerNorm transforms a vector in representation space.

\item Present LayerNorm in action with experiments depicting its importance in stabilizing the residual stream vectors. 

\item Show that removing the component along the uniform vector is a redundant step in LayerNorm at inference time. These findings advocate for the adoption of RMSNorm over LayerNorm in LLMs.

\end{enumerate}

\section{Re-Introducing Layer Normalization}\label{sec:reintroducing-layernom}

Let $\bm{x} \in \mathbb{R}^d$ represent an intermediate hidden representation in a transformer-based model, on which the LayerNorm operation is applied. Then the following steps summarize the layer normalization operation:

\begin{enumerate}
    \item Calculate the mean, $\mu = \frac{1}{d} \sum^d_{i=1} x_i$, and standard deviation, $\sigma = \sqrt{\frac{1}{d} \sum^d_{i=1} (x_i - \mu)^2} $, where $x_i$ are the components of $\bm{x}$
    \item Standardize the components of the hidden vector $\bm{x}$ to get $\bm{y}$, such that $y_i = \frac{x_i - \mu}{\sigma}$, $\forall x_i \in \bm{x}$
    \item Scale and shift to get $\bm{z} = \bm{\alpha} \odot \bm{y} + \bm{\beta}$, where $\odot$ represents element-wise multiplication of two vectors, and $\bm{\alpha}$, $\bm{\beta}$ are scaling and shifting vector parameters learned during training.
\end{enumerate}

In this paper, we refer to the combination of step-1 and step-2 as the \textit{"standardization step"}, whereas step-3 is referred to as the \textit{"scale-and-shift"} step. The removal of the mean of components from each component in step-1 is referred to as the \textit{"mean subtraction"} step in this paper. For an overview of the computations happening within a decoder layer in modern LLMs, we refer the reader to section \ref{subsection:detailed-layer}.

While the original definition of LayerNorm gives us an operational description, it tells us very little about the global properties of the resulting vectors. For example, with the steps described above, we get very little insight into the norm or the orientation of the standardized vector ($\bm{y}$). All we know is that the components of the vector $\bm{y}$ are standardized to have zero mean and unit variance within the vector. But how is this standardized vector oriented compared to the original vector and what are its norms? Such an explanation was also absent in the original formulation of layer normalization \citep{layernorm}.

Initial explorations into the geometry of layer normalization by \citet{meaning-layernorm} demonstrated that $\bm{y}$ maintains a norm of $\sqrt{d}$ and is oriented orthogonal to the constant vector $\boldsymbol{1} = [1, 1, 1, 1, \cdots, 1]^T \in \mathbb{R}^d$, where $d$ represents the dimensionality of the vector space of hidden vectors. Furthermore, a popular variant of LayerNorm, known as RMSNorm \citep{rmsnorm}, which omits the ``mean subtraction" in the standardization step, has come out as a viable alternative to LayerNorm. In this paper, we expand upon these foundational studies by presenting a novel interpretation of the global effects of LayerNorm on a vector, which we argue provides a more intuitive and informative description of how LayerNorm modifies a hidden vector in representation space. Additionally, we analyze the representations of models using RMSNorm, showing that despite omitting the "mean subtraction step", RMSNorm produces hidden representations with similar orientations.

To understand the global effect of LayerNorm on a vector $\bm{x}$, we need to represent the standardization steps in terms of the vector $\bm{x}$ and not its components. A neat way to write the mean of the components of a vector $\bm{x}$ is:

\begin{equation}\label{eq:mean}
    \mu = \frac{1}{d} \boldsymbol{1}^T \bm{x}
\end{equation}

where $\boldsymbol{1} = [1, 1, 1, 1, \cdots 1]^T$ such that $\boldsymbol{1} \in \mathbb{R}^d$, which we refer to as the \textit{uniform vector} in this paper. Using the scalar mean value calculated above, we define the \textit{mean vector} as 

\begin{equation}\label{eq:mean_vector}
    \bm{\mu} = \mu\boldsymbol{1}
\end{equation}

where $\mu$ is a scalar mean as calculated in equation \ref{eq:mean}. Thus, the mean vector is a vector with each component equal to the mean of the components of vector $\bm{x}$. Using this, we can also rewrite the standard deviation of the components of $\bm{x}$ as:

\begin{equation}\label{eq:std}
    \sigma = \sqrt{\frac{1}{d} (\bm{x} - \bm{\mu})^T(\bm{x} - \bm{\mu})}
\end{equation}

Once we do this, we can now write the standardize step in the LayerNorm algorithm (step-2) in vector form as follows:

\begin{equation}\label{eq:standardization}
    \bm{y} = \frac{1}{\sigma} (\bm{x} - \bm{\mu})
\end{equation}

In the above equation, we are assuming that the standard deviation is non-zero. As can be seen in equation \ref{eq:std}, the standard deviation is zero only when $\bm{x} = \bm{\mu}$, which happens when all components of the hidden vector are equal. In practice, the authors of LayerNorm add an error term in the denominator to prevent this from happening. In the discussion that follows, we will assume that $\bm{x} \neq \bm{\mu}$, which is the same as saying that all components of $\bm{x}$ are not equal. This assumption does not lead to any loss of generality, as will be evident in later discussions.  If the standard deviation is zero or $\bm{x} = \bm{\mu}$, then LayerNorm outputs a vector with all its components equal to zero. 



\subsection{The Uniform Vector and the Mean Vector}
The vector definition of layer normalization in equation \ref{eq:standardization} requires us to define two new vectors: the \textit{uniform vector} and the \textit{mean vector}. Since these are non-standard vectors, it is important to understand the properties of these vectors.

The \textit{uniform vector} or $\boldsymbol{1} = [1, 1, 1, 1, \cdots 1]^T$, is called so because all its components are equal or uniform and set to 1. This vector plays a very important role in the formulation and understanding of layer normalization. An important thing to note here is the norm of the uniform vector: $\|\boldsymbol{1}\|_2 = \sqrt{\boldsymbol{1}^T\boldsymbol{1}} = \sqrt{d}$.


The \textit{mean vector} in the context of LayerNorm is defined in a non-traditional manner. Traditionally, a mean vector is the sum of a few vectors. But in our definition, the mean vector, $\bm{\mu} = \mu\boldsymbol{1}$, is a scaled version of the uniform vector. It is scaled by a value that is the mean or the average of all the components of the hidden vector and is oriented in the direction of the uniform vector. An interesting property of the mean vector is its norm. Let's define $\theta_{x1}$ as the angle between vectors $\bm{x}$ and the uniform vector. Then taking the L2 norm of the mean vector:

\begin{equation}\label{eq:mean-vector-norm}
\begin{split}
\| \bm{\mu}\|_2 &= \|\mu \boldsymbol{1}\|_2 = \mu \|\boldsymbol{1}\|_2 = \left( \frac{1}{d} \boldsymbol{1}^T \bm{x}\right) \|\boldsymbol{1}\|_2 \\
&= \frac{1}{d}\|\boldsymbol{1}\|_2^2 \|\bm{x}\|_2 \cos\theta_{x1} = \|\bm{x}\|_2 \cos\theta_{x1}
\end{split}
\end{equation}

Here we replace the formula for the mean from equation \ref{eq:mean}, expand the inner product between the uniform vector and $\bm{x}$ in terms of their norms and the angle between them, and use the fact that $\|\boldsymbol{1}\|_2 = \sqrt{d}$.

The norm of the \textit{mean vector} gives us very important insights about what's going on in the layer normalization process. Equation \ref{eq:mean-vector-norm} shows that the norm of the mean vector is nothing but \textbf{the projection of the underlying vector $\bm{x}$ along the uniform vector}. By definition, the mean vector, $\bm{\mu} = \mu\boldsymbol{1}$, is oriented along the direction of the uniform vector. In other words, given that $\theta$ is the angle between vectors $\bm{x}$ and the uniform vector,

\begin{equation}
    \begin{split}
\| \bm{\mu}\|_2 = \|\bm{x}\|_2 \cos\theta_{x1} = \frac{\bm{x} \cdot \boldsymbol{1}}{\|\boldsymbol{1}\|_2} = \bm{x} \cdot \hat{\boldsymbol{1}}
    \end{split}
\end{equation}

and 

\vspace{-20pt}
\begin{equation}
    \begin{split}
 \bm{\mu}\ = (\bm{x}.\hat{\bm{1}}) \hat{\bm{1}}
    \end{split}
\end{equation}

Note that $\hat{\bm{1}}$ is the unit vector corresponding to the uniform vector $\bm{1}$. To the best of our knowledge, the geometrical meaning of the mean vector has not been discussed in literature before our work. In the next section, we incorporate this information to explain the LayerNorm process, giving a very intuitive and informative description of the process.


\subsection{Explanation of Layer Normalization}\label{sec:layernom-explanation}

If we go through the steps of layer normalization as discussed in the beginning of section \ref{sec:reintroducing-layernom}, the component-wise subtraction of the mean of all components of a vector can now be written completely in terms of the vector $\bm{x}$. We define $\bm{x^{\boldsymbol{1}}_{\perp}}$ as the component of $\bm{x}$ orthogonal to the uniform vector as follows:

\begin{equation}
    \bm{x^{\boldsymbol{1}}_{\perp}} = \bm{x} - \bm{\mu} = \bm{x} - (\bm{x}.\hat{\bm{1}}) \hat{\bm{1}}
\end{equation}

\textbf{Thus, subtracting the mean of the components of a vector is the same as the removal of the projection of the vector along the uniform vector.} The complete standardization step of the layer normalization operation can now be written as:


\begin{equation}\label{eq:redef}
    \bm{y} =\sqrt{d} \frac{\bm{x^{\boldsymbol{1}}_{\perp}}}{\|\bm{x^{\boldsymbol{1}}_{\perp}}\|_2}
\end{equation}

where $d$ is the dimensionality of the vector being normalized. This shows that \textbf{layer normalization can be simply defined as the normalization of the component of a vector orthogonal to the uniform vector, accompanied by a scaling factor}. With this, we present the simple and elegant definition of the layer normalization process with a deep geometric meaning. The most intuitive recipe of the layer normalization process is shown in Algorithm~\ref{algorithm:x}.

\begin{algorithm}
\caption{\textbf{: The Standardization Step in Layer Nor}}
\label{algorithm:x}
\begin{algorithmic}[1]
 \State  Throw away the component of a vector along the uniform vector, $\boldsymbol{1} = [1, 1, 1, 1, \cdots 1]^T$
 \State  Normalize the remaining vector
 \State  Scale the resulting vector by $\sqrt{d}$, where $d$ is the dimensionality of the vector space
\end{algorithmic}
\end{algorithm}

Equation \ref{eq:redef} also shows that the norm of the standardized vector is $\sqrt{d}$ and is oriented orthogonal to the direction of the uniform vector, which means it exists in a subspace orthogonal to the uniform vector.

\subsection{The Irreversibility of Layer Normalization}\label{sec:irrerverible}
 The idea of reversibility is discussed briefly while introducing batch normalization \citep{batchnorm}, which normalizes each feature dimension in the input vector $\bm{x}$ independently by calculating the mean and standard deviation statistics over a large sample of such vectors. Specifically, let $\bm{x}^1, \bm{x}^2, \dots \bm{x}^b$ be a set of $b$ vectors in the training set, each vector of dimension $d$. Then, both the batch normalization and layer normalization processes for each component can be represented in the following two-step process:

\begin{align}
    y_i = \frac{x_i -\mathbb{E}[x_i]}{\sqrt{Var(x_i)}}  \hspace{4pt} \text{(standardization)}\\
    z_i = \alpha_ix_i + \beta_i \hspace{4pt} \text{(scale-and-shift)}
\end{align}

The key difference between batch normalization and layer normalization lies in how the expectation values and variance are calculated for each component. In the case of batch normalization, the mean and variance for each dimension are calculated over the training set. This means that $\mathbb{E}[x_i]$ and $Var(x_i)$ are the same for each vector $\bm{x}^j \in \{\bm{x}^1, \bm{x}^2, \dots \bm{x}^b\}$, but different for each component of the vector. In total, for a $d$-dimensional hidden vector space, there are $2d$ statistics in batch normalization, $d$ for the mean and $d$ for variance. Due to this, the information lost during the standardization step in batch normalization, if important, can be recovered in the scale-and-shift step by simply learning $\alpha_i = \sqrt{Var(x_i)}$ and $\beta_i = \mathbb{E}[x_i]$, since we have $2d$ learnable parameters in batch normalization. The scale-and-shift step was a very conscious design choice of the inventors of batch normalization to allow for batch normalization to represent an identity transformation if the original hidden representations were optimal for the network. 

A subtle but very important difference with layer normalization is how the expectation and variances are calculated for each component. For layer normalization, these statistics are calculated for each vector independently. $\mathbb{E}[x_i]$ and $Var(x_i)$ in layer normalization are the same for each component $x_i$ within a vector but are different for different vectors. This means that two variables are created in LayerNorm per hidden vector. As LLMs will normalize way more than $d$ vectors throughout the course of their training, thus creating more than $2d$ statistics, the number of statistic variables used in LayerNorm is very high. Thus, the information lost during layer normalization cannot be recovered by learning just $2d$ components of the $\bm{\alpha}$ and $\bm{\beta}$ learnable parameters. This shows that \textbf{information once lost during layer normalization cannot be recovered during the scale-and-shift step, making the layer normalization process irreversible}. 

\setlength{\abovecaptionskip}{25pt}
\begin{figure*}[t]
    \centering
    \begin{subfigure}[b]{0.32\textwidth}
        \centering
        \includegraphics[width=\textwidth]{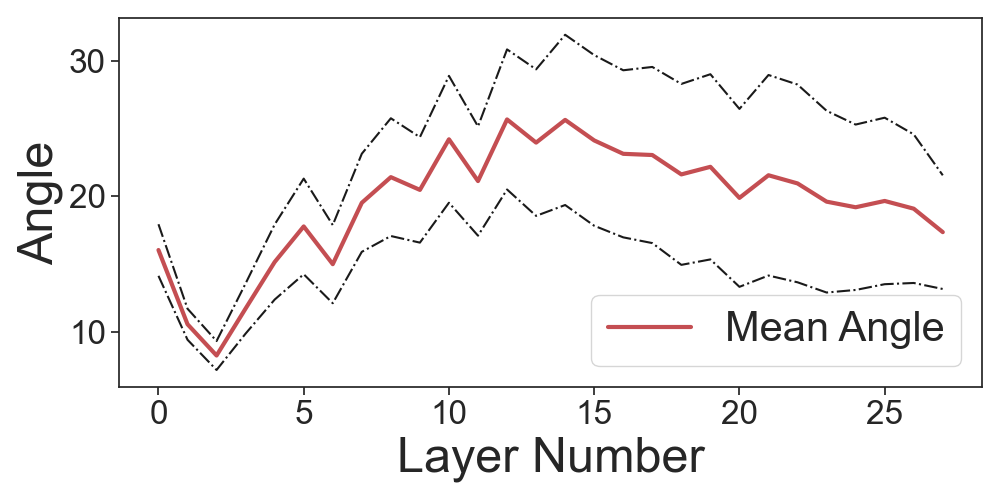}
        \caption{GPT-J}
        \label{fig:rotGPTJln1}
    \end{subfigure}
    \begin{subfigure}[b]{0.32\textwidth}
        \centering
        \includegraphics[width=\textwidth]{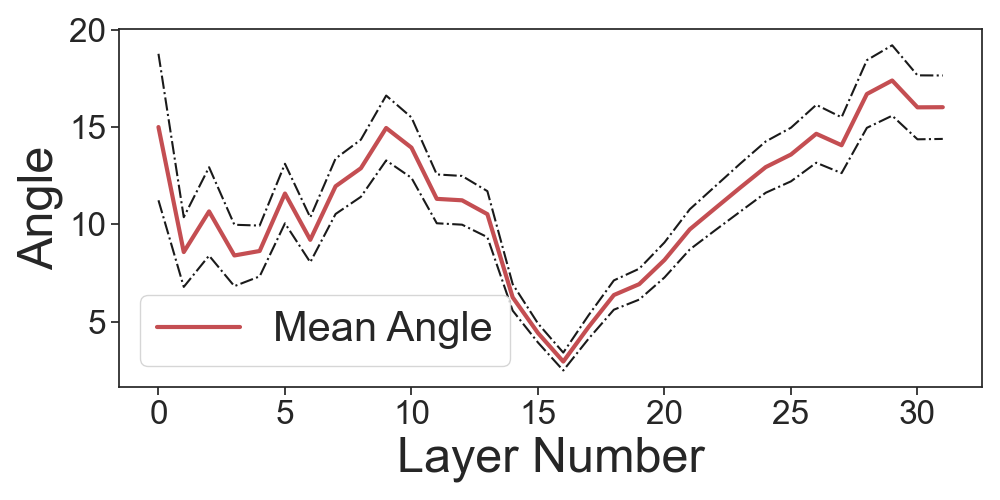}
        \caption{Pythia 6.9}
        \label{fig:rot6.9ln1}
    \end{subfigure}
    \begin{subfigure}[b]{0.32\textwidth}
        \centering
        \includegraphics[width=\textwidth]{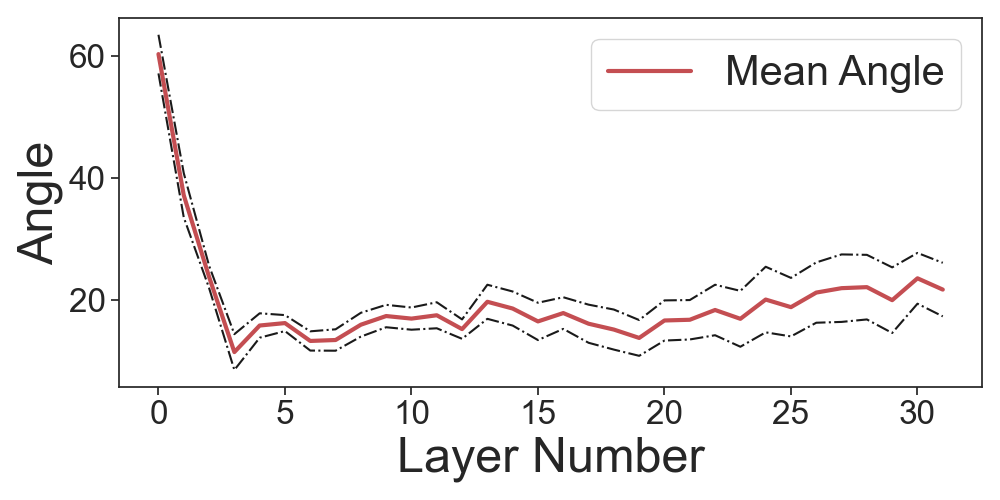}
        \caption{Llama-3}
        \label{fig:rotLLAMA3ln1}
    \end{subfigure}
    \caption{Rotation angle (in degrees) between the hidden vectors and Post-LN1 vectors across all layers for GPT-J, Pythia 6.9, Llama-3}
    \label{fig:rotationanglefigure}
\end{figure*}

\begin{table}
  \centering
  \small
  \begin{tabular}{lcccc}
    \toprule
        Model & Model &  Num    & Num        & Norm. \\
      Name   & Dim (d)   &  Params  &    Layers  & Type \\
    \midrule
    GPT-2 XL      & 1600 & 1.5B & 48 & Layer \\
    GPT-Neo 1.3B  & 2048 & 1.3B & 24 & Layer \\
    Pythia-1.4B   & 2048 & 1.4B & 24 & Layer \\
    GPT-J 6B      & 4096 & 6.0B & 28 & Layer \\
    Pythia-6.9B   & 4096 & 6.9B & 32 & Layer\\
    Llama-2-7B    & 4096 & 7.0B & 32 & RMS \\
    Llama-3-8B    & 4096 & 8.0B & 32 & RMS \\
    \bottomrule
  \end{tabular}
  \caption{List of models used in experiments. }
  \label{tab:model_details}
\end{table}

\section{Experiments}
After a detailed theoretical analysis of the effects of layer normalization on a vector, we next study how LayerNorm affects the hidden representations of LLMs in practice. While prior works have studied the effects of LayerNorm on downstream model performance and training convergence \citep{layernorm, understanding-layernorm, rmsnorm}, there has been a surprising gap in the literature about studying the effect of LayerNorm on the internal representations of a model at inference time. To do so, we pass one million tokens from Wikipedia articles through various models and study how LayerNorm modifies a hidden vector in the representation space.



\setlength{\abovecaptionskip}{5pt}
\setlength{\belowcaptionskip}{0pt}
\begin{figure*}[t]
    \centering
    \begin{subfigure}[b]{0.32\textwidth}
        \centering
        \includegraphics[width=\textwidth]{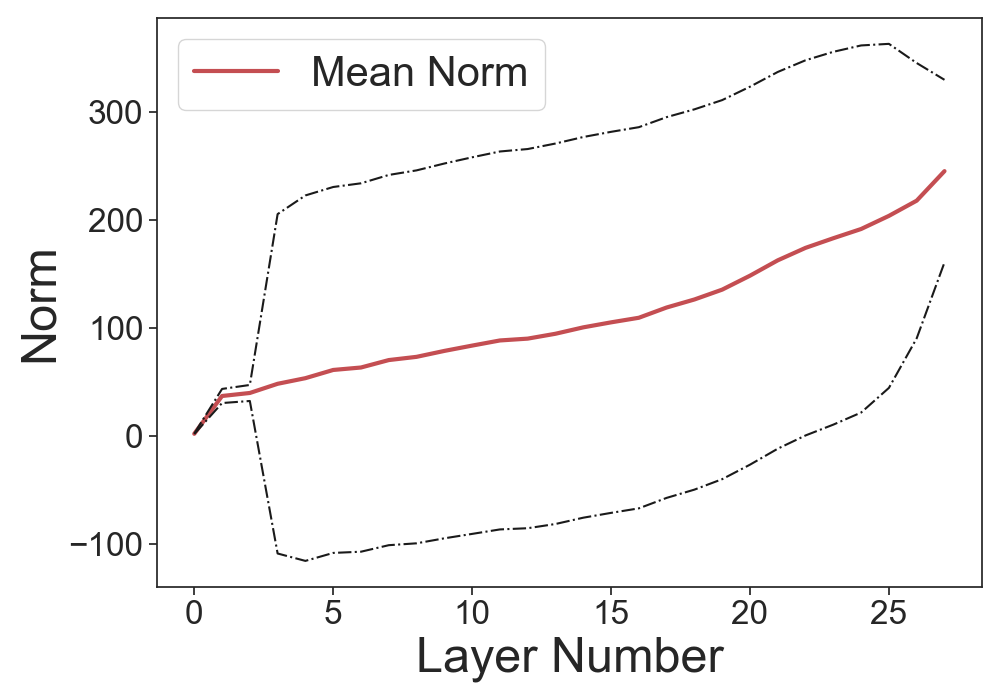}
        \caption{Residual Stream Pre-LN1 GPT-J}
        \label{fig:residual-norm-prelnGPTJ}
    \end{subfigure}
    \centering
    \begin{subfigure}[b]{0.32\textwidth}
        \centering
        \includegraphics[width=\textwidth]{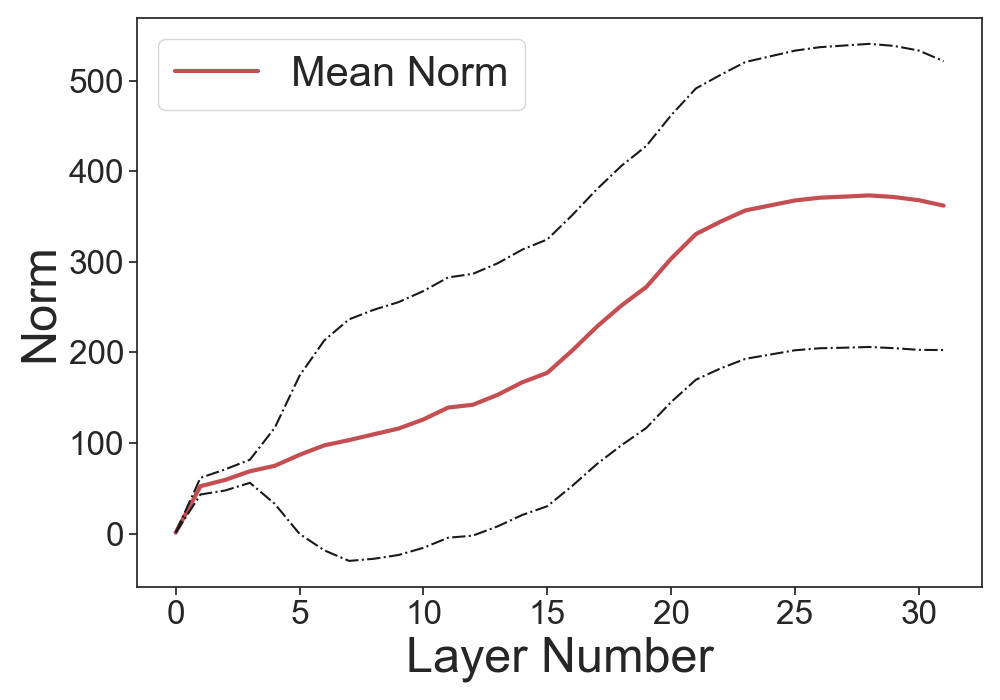}
        \caption{Residual Stream Pre-LN1 Pythia 6.9}
        \label{fig:residual-norm-preln6.9}
    \end{subfigure}
    \centering
    \begin{subfigure}[b]{0.32\textwidth}
        \centering
        \includegraphics[width=\textwidth]{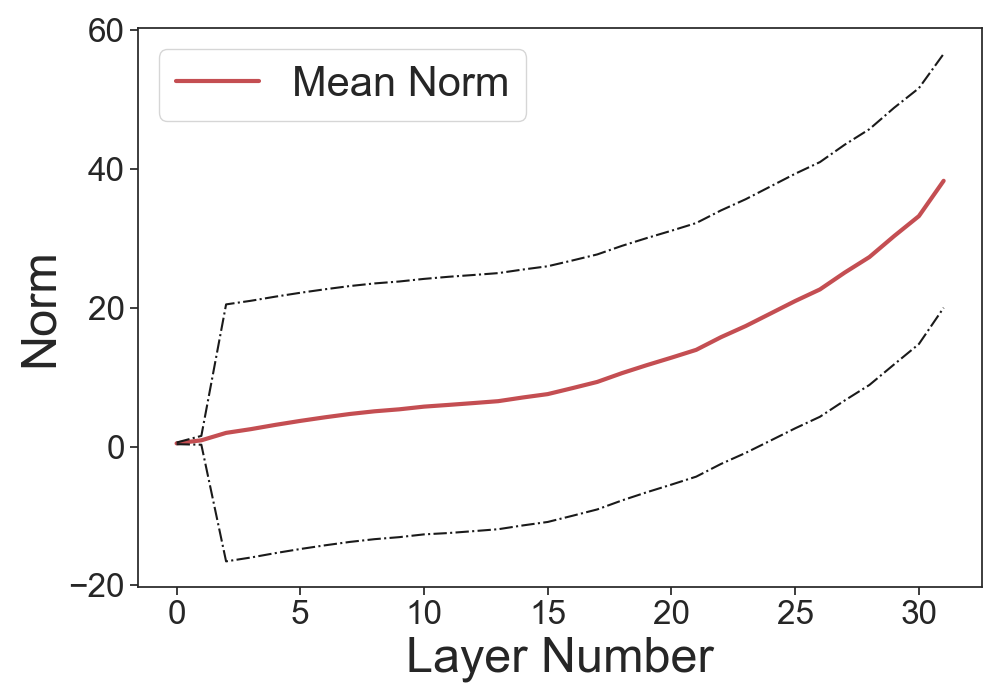}
        \caption{Residual Stream Pre-LN1 Llama-3}
        \label{fig:residual-norm-prelnLLAMA3}
    \end{subfigure}
    \hfill
    \begin{subfigure}[b]{0.32\textwidth}
        \centering
        \includegraphics[width=\textwidth]{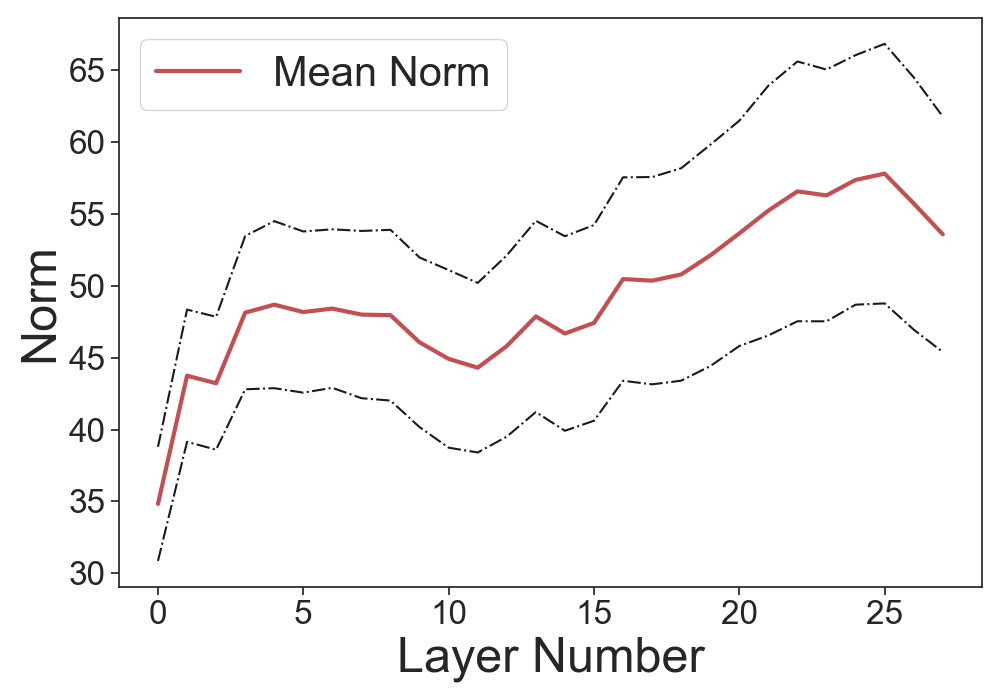}
        \caption{Residual Stream Post-LN1 GPT-J}
        \label{fig:residual-norm-postlngptj}
    \end{subfigure}
    \begin{subfigure}[b]{0.32\textwidth}
        \centering
        \includegraphics[width=\textwidth]{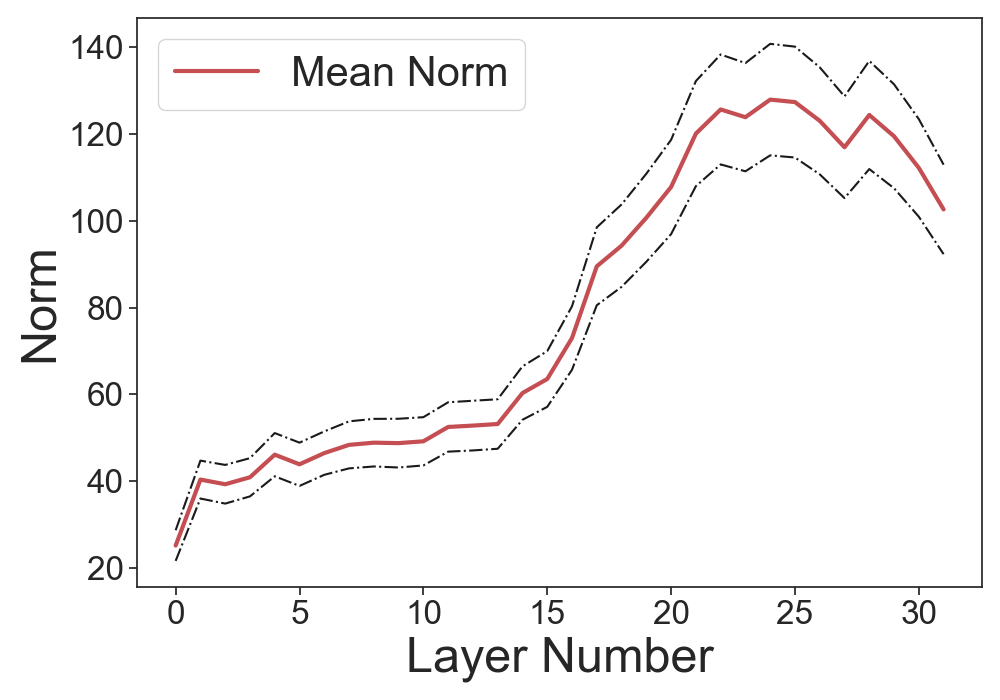}
        \caption{Residual Stream Post-LN1 Pythia 6.9}
        \label{fig:residual-norm-postln6.9}
    \end{subfigure}
    \begin{subfigure}[b]{0.32\textwidth}
        \centering
        \includegraphics[width=\textwidth]{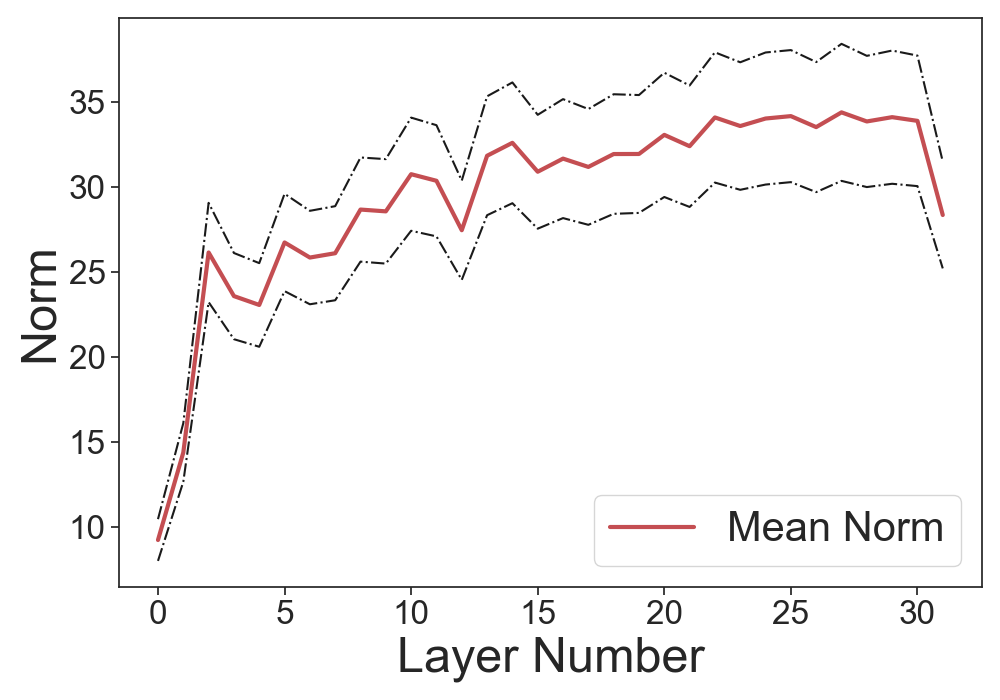}
        \caption{Residual Stream Post-LN1 Llama-3}
        \label{fig:residual-norm-postlnLLAMA3}
    \end{subfigure}
    \caption{This figure shows the growing norm of the residual stream or hidden vectors at each layer (a-c) and how LayerNorm and RMSNorm regulate the growing norms (d-f) for GPTJ 6B, Pythia 6.9B and Llama3 8B. The dashed lines in the line plots represent one standard deviation.}
    \label{fig:residual-norm-ln-effect}
\end{figure*}

\paragraph{Methods and Models} 
We study the impact of LayerNorm on the hidden representations of 7 decoder-only LLMs across two size categories. The models used are listed in Table \ref{tab:model_details}. GPT2XL, GPT-Neo-1.3B, and Pythia-1.4B represent the small LLM size category, whereas GPT-J-6B, Pythia-6.9B, Llama-2-7B, and Llama-3-8B represent the medium LLM size category. We pass one million tokens from Wikipedia articles through each model and capture the hidden representations for all tokens before and after each normalization layer. This is done for each layer inside the model, which creates many terabytes of data to be analyzed. Norms were calculated using the L2 norm, and angle differences in degrees were determined through cosine similarity calculations to measure the orientation changes induced by normalization.

As outlined in Table \ref{tab:model_details}, we also use two types of normalization methods. Models like GPT-2 XL, GPT-Neo 1.3B, Pythia-1.4B, GPT-J 6B, and Pythia-6.9B employ LayerNorm \cite{layernorm}, whereas Llama-2-7B and Llama-3-8B utilize RMSNorm \cite{rmsnorm}. The crucial difference between LayerNorm and RMSNorm is the subtraction of the mean vector (Figure \ref{fig:lndiagram}). In this paper, we study how LayerNorm and RMSNorm operate on the hidden representations.




\subsection{Norm Stabilization}\label{sec:norm-stabilization}
As a token representation passes through the different layers inside an LLM, the hidden representations accumulate due to residual connections \cite{resnet}. These hidden representations get added to the residual stream at each layer as shown in equation \ref{eq:resiudal-summation} and can cause the norm of the hidden vectors to grow. In fact, the norm of the hidden vectors at each layer grows disproportionately with large standard deviations, as can be seen in Figure \ref{fig:residual-norm-ln-effect} (a-c). The growing norm of the hidden vectors is extreme for some models more than others, as is seen for GPT-J and Pythia-6.9B. The growing norm of hidden representations for the remaining models from Table \ref{tab:model_details} and the effect of layer normalization can be seen in Figure \ref{fig:residual-norm-ln-effect_rest} (appendix).

The significant role of LayerNorm and RMSNorm in stabilizing the growing norm of the hidden vectors at each layer can be seen in Figure \ref{fig:residual-norm-ln-effect} (d-f). The standardization step in layer normalization first modifies the norm of each vector to $\sqrt{d}$, where $d$ is the dimensionality of the representation space. The norms of the standardized vectors are further modified marginally by the scale-and-shift steps. Layer normalization affects both the mean and the spread of the norms, where the standard deviations in the norm are reduced by factors between 10-100 depending on the model (Table \ref{tab:norm_table} in appendix). These results show the crucial role that LayerNorm and RMSNorm play in stabilizing the intermediate hidden vectors in practice.



\subsection{Rotation}
While stabilizing the growing norm of hidden vectors is a major function of layer normalization, it also ends up rotating the hidden vectors. Figure \ref{fig:rotationanglefigure} shows the angle between the original hidden vector and the post-normalization vector. As can be seen in Figure \ref{fig:rotationanglefigure} (and in Figure \ref{fig:rotationanglefigurerest} for the rest of the models in the appendix), each layer is responsible for a fixed, non-trivial amount of rotation of hidden vectors in the representation space. Mean rotation angles in degrees between hidden and post-layer normalization vectors (LN1 and LN2) across all layers for various models are shown in Table \ref{tab:rotation-table} in the appendix. Thus, both LayerNorm and RMSNorm additionally rotate hidden vectors in representation space in addition to stabilizing the norms of the vectors.

\begin{figure*}[t]
    \centering
    \begin{subfigure}[b]{0.32\textwidth}
        \centering
        \includegraphics[width=\textwidth]{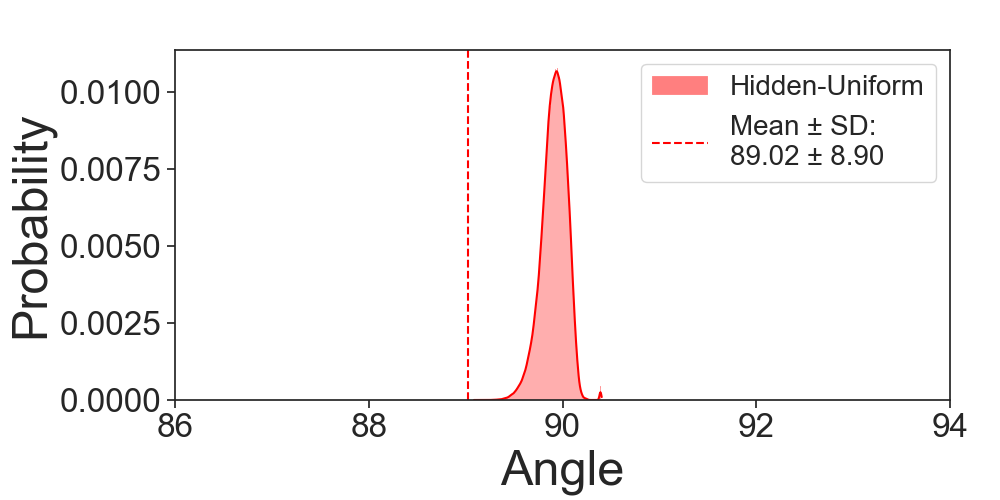}
        \caption{Hidden GPT-J}
    \end{subfigure}
    \begin{subfigure}[b]{0.32\textwidth}
        \centering
        \includegraphics[width=\textwidth]{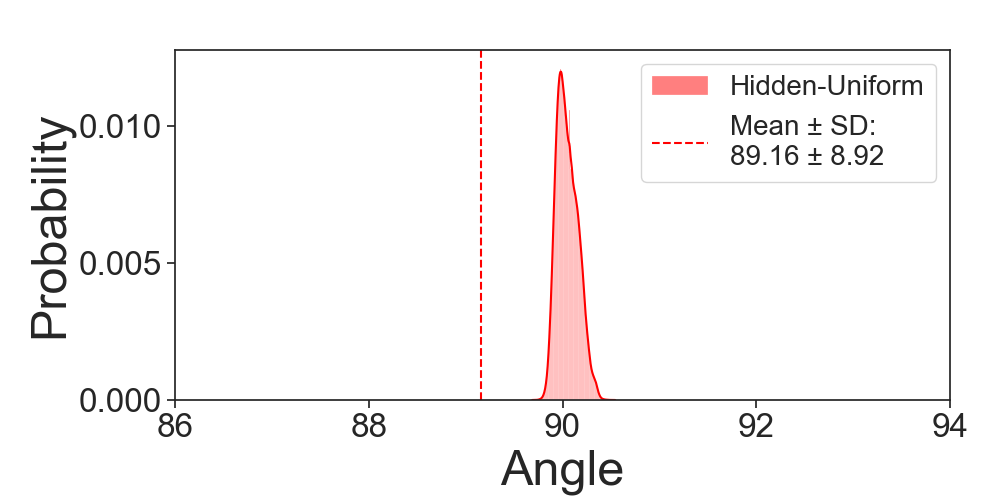}
        \caption{Hidden Pythia 6.9}
  
    \end{subfigure}
    \begin{subfigure}[b]{0.32\textwidth}
        \centering
        \includegraphics[width=\textwidth]{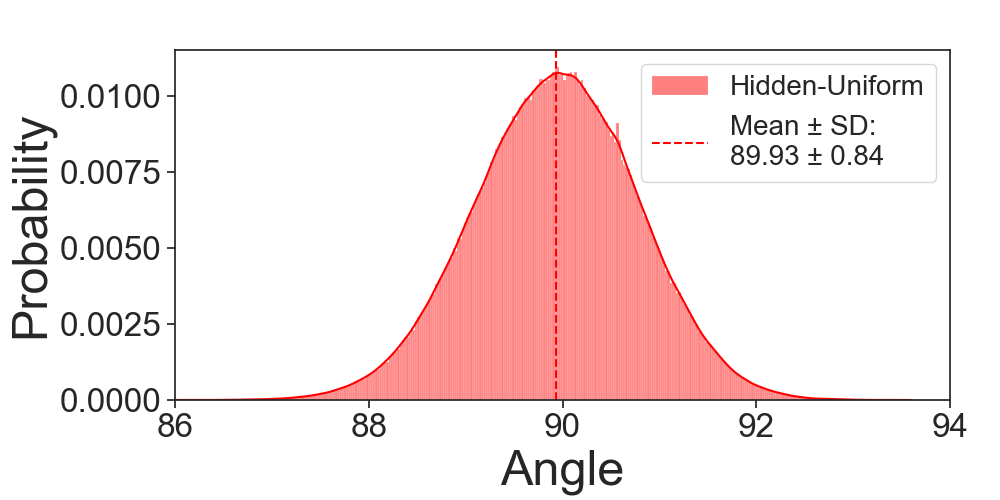}
        \caption{Hidden Llama-3}
      
    \end{subfigure}
    \hfill
    \begin{subfigure}[b]{0.32\textwidth}
        \centering
        \includegraphics[width=\textwidth]{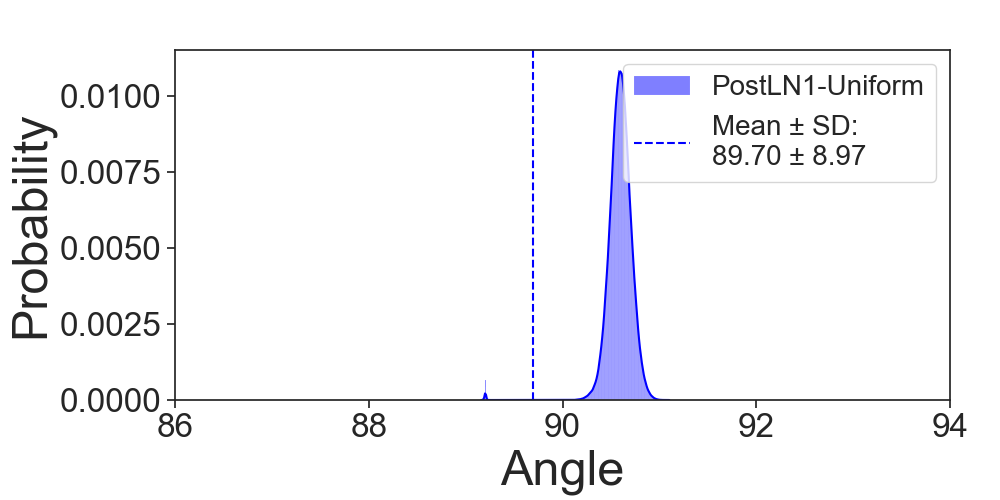}
        \caption{Post-LN1 GPTJ}
       
    \end{subfigure}
    \begin{subfigure}[b]{0.32\textwidth}
        \centering
        \includegraphics[width=\textwidth]{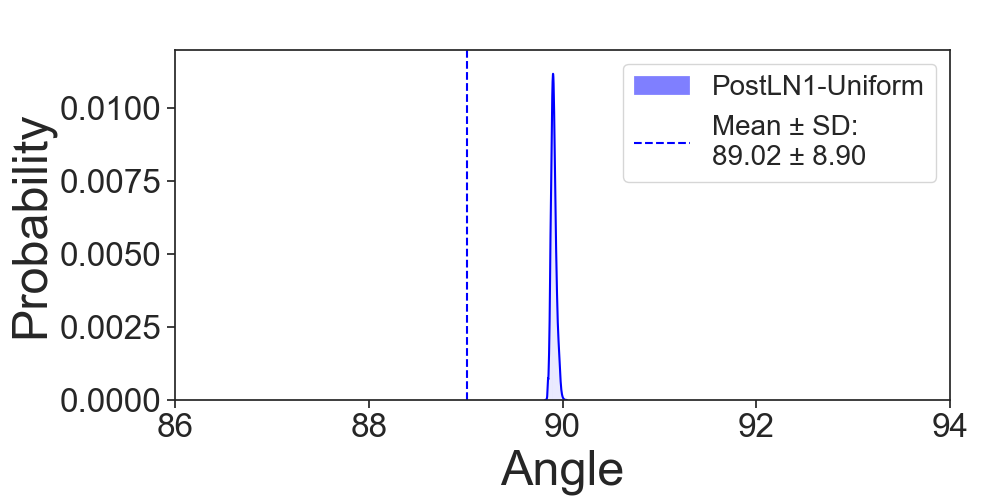}
        \caption{Post-LN1 Pythia 6.9}
       
    \end{subfigure}
    \begin{subfigure}[b]{0.32\textwidth}
        \centering
       \includegraphics[width=\textwidth]{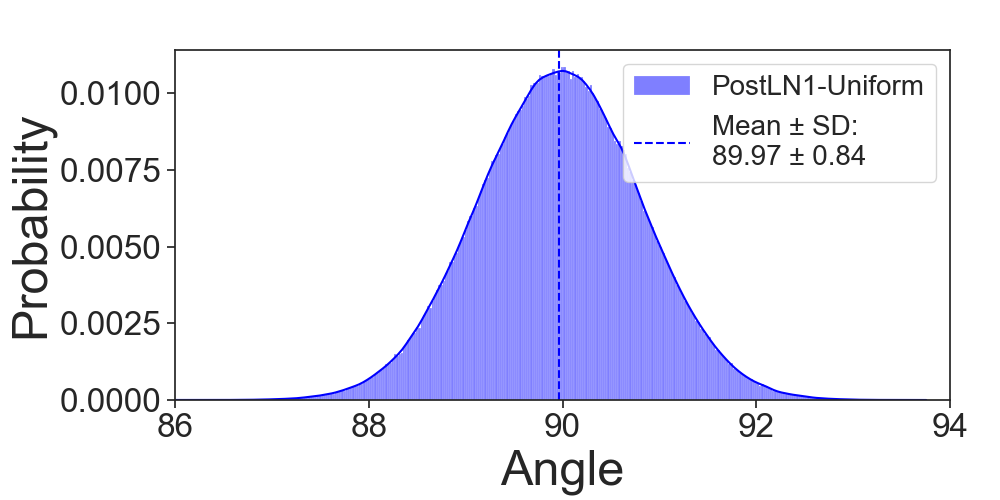}
        \caption{Post-LN1 Llama-3}
       
    \end{subfigure}

    \caption{Distribution of angles (in degrees) between Hidden vectors (a-c) and post-normalization vectors (d-f) with the \textit{uniform vector} for GPT-J, Pythia 6.9, Llama-3 for a randomly selected layer (Layer 24). The results are independent of the choice of layers. }
    \label{fig:angle-diff-main-paper}
\end{figure*}

\section{LayerNorm versus RMSNorm}\label{sec:layernorm_vs_rms}

Apart from norm stabilization and rotation, a critical aspect of LayerNorm is its ability to orient the hidden vectors orthogonal to the uniform vector, as discussed in section \ref{sec:reintroducing-layernom}. In section \ref{sec:layernom-explanation}, we see that the definition of LayerNorm is innately linked with the uniform vector, possibly unintended by the original authors. Since the information along the uniform vector is being removed irreversibly during layer normalization (section \ref{sec:irrerverible}), the layer normalization process implicitly assumes that the information along the uniform vector is either not important or that the model should not store information along that direction. However, two randomly chosen vectors in high-dimensional spaces are nearly orthogonal to each other with a very small spread. Since the ``mean subtraction" step in LayerNorm means removing the component along the arbitrarily chosen direction of the uniform vector, the need for such a step would be justified if the hidden representations created by the model had unnaturally significant components along this vector. However, there seems to be no justification given in the literature for the mean subtraction step, including the original paper \cite{layernorm}. 

RMSNorm, on the other hand, does not perform this step and simply normalizes the existing vector \cite{rmsnorm}. The Llama model series \citep{llama, llama2} is the most prominent family of LLMs that use RMSNorm. The simple fact that Llama models achieve state-of-the-art results across multiple measures \citep{llama3} shows that using RMSNorm instead of LayerNorm does not hurt performance. This provides a strong motivation to explore the need for the ``mean subtraction" step in LayerNorm.  Thus, we ask the question - do intermediate hidden representations have a non-trivial component along the uniform vector that justifies its removal?

We study this question at inference time. To answer this question, we measure the angle between the hidden vector and the uniform vector just before and after the normalization operations for LayerNorm and RMSNorm-based models. To justify the mean subtraction step, one of the following two scenarios should be true: 
\begin{itemize}
    \item Scenario-1: In LayerNorm-based models, the intermediate hidden vectors have a large component along the uniform vector pre-normalization, which gets removed post-normalization. 
    \item Scenario-2: In RMSNorm-based models, which are trained without the mean subtraction step, the hidden vectors consistently have a large component along the uniform vector.
\end{itemize}


The results are shown in Figure \ref{fig:angle-diff-main-paper} for GPT-J (6B), Pythia (6.9B), and Llama-3 (8B) and Figure \ref{fig:angle-diff-main-paperREST} for the remaining models. We see that for all models in Figure \ref{fig:angle-diff-main-paper}, the angle between the hidden vectors and the uniform vectors is 90 degrees on average even before normalization, with a very small spread. This remains true post-normalization as well for all three models. This result is surprising because the hidden representations for LayerNorm-based models (GPT-J and Pythia) are themselves orthogonal to the uniform vector even before the LayerNorm operation. This is true for 4 out of the 5 LayerNorm-based models studied in this paper except for GPT2-XL (Figure \ref{fig:angle-diff-main-paperREST} in the appendix). This shows that the ``mean subtracting" operation of LayerNorm is redundant during inference, as there is nothing to remove. 




For RMNSNorm-based LLMs, the result for Llama-3 (8B) in Figure \ref{fig:angle-diff-main-paper} represents scenraio -2. We see that even for Llama-3 (8B) trained using RMSNorm, the hidden representations before normalization are also on average orthogonal to the uniform vector with a very small spread. In fact, they are more orthogonal to the uniform vector than for LayerNorm-based models, with a much smaller variance. After RMSNorm, the hidden representations continue to be orthogonal to the uniform vector. This is also true for Llama-2 (7B) (Figure \ref{fig:angle-diff-main-paperREST} in the appendix). This shows that even without the ``mean subtraction" step during training as in RMSNorm, the hidden vectors operate orthogonal to the uniform vector, underscoring the redundance of the ``mean subtraction" step. 


These results show that the ``mean subtracting" operation in LayerNorm is redundant as the model naturally aligns its representations orthogonal to the uniform vector as expected in such high-dimensional spaces. In fact, in hindsight, evidence of having large components along the uniform vector would have been more surprising and indicated something fundamentally important about the specific vector. With this, we provide the first mechanistic evidence that the \textbf{``mean subtraction" step in LayerNorm is dispensable}, as we do not find hidden vectors having large components along the uniform vector. This line of investigation was made possible by our novel geometrical interpretation of LayerNorm (section \ref{sec:layernom-explanation}), which presented the global implications of the mean subtraction step in LayerNorm - removing the component of intermediate hidden representations along the uniform vector. Due to these results, we advocate for using RMSNorm over LayerNorm, which is also computationally more efficient and leads to comparable downstream performance.

\section{Conclusion}
We present a detailed theoretical and empirical analysis of layer normalization on hidden vectors in representation space, with a focus on the global effects of the LayerNorm operation on a vector. We first show that the LayerNorm operation can be understood in three simple steps: removing the component of a vector along the uniform vector ($\boldsymbol{1} = [1, 1, 1, 1, \cdots, 1]^T \in \mathbb{R}^d$), normalizing the remaining vector, and scaling the resultant vector by $\sqrt{d}$. We then show that LayerNorm is an irreversible process—information along the uniform vector is removed during LayerNorm and cannot be recovered using the learnable parameters available in the formulation. This is in contrast with BatchNorm, where the network has the option of learning an identity operation. We then empirically illustrate how LayerNorm regulates the norm and the orientation of hidden vectors during model inference. Finally, we show that the ``mean subtraction" operation in LayerNorm is dispensable both during inference and for training, as shown by internal representations of RMSNorm-based Llama models. 


\section{Limitations}
This study provides valuable insights into the theoretical and empirical effects of layer normalization. Although the study empirically tests 7 models across two sizes and multiple language model families, the size of the models tested is restricted to 8 billion parameters. Since we wanted our results to have a large enough sample size, we chose to store hidden representations of 1 million tokens. Even for a small model like GPT2-XL, this requires 2TB of data, and for larger models like Llama3-8B, it requires storing approximately 4TB of data. Storing such large amounts of hidden representations pushed the capacity of the computational resources available to us. Because of this restriction, we leave testing our findings for larger models to future works and groups that have access to larger amounts of computing. 

\bibliography{custom}

\appendix

\section{Appendix}
\label{sec:appendix}

\subsection{Computations in Decoder-only LLMs}\label{subsection:detailed-layer}
\begin{figure}[t]
        \centering
        \includegraphics[width=0.4\columnwidth]{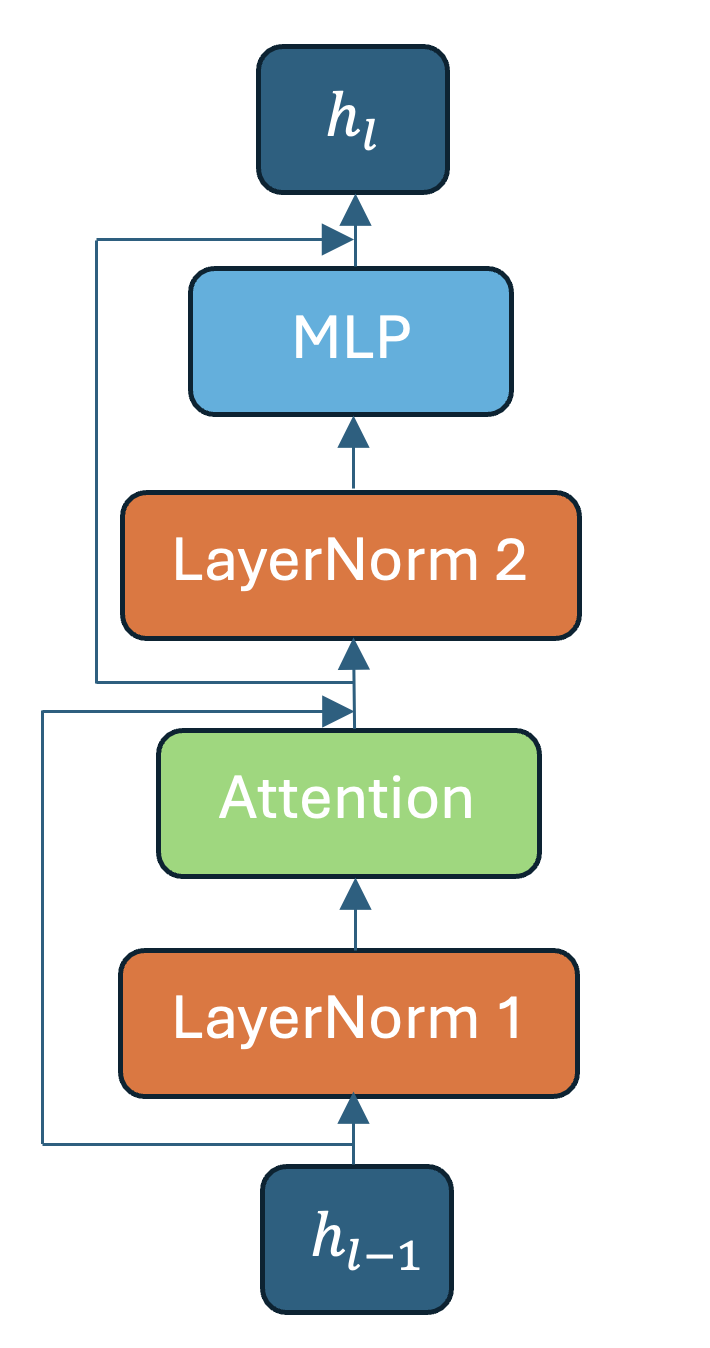}
    \caption{A high-level diagram representing the computation within one decoder block of an LLM.}
    \label{fig:transformer}
\end{figure}

In this paper, we study the hidden representations of modern decoder-only large language models (LLMs). Let $h^l$ represent the intermediate hidden state vectors between each decoder layer. 
As depicted in Figure \ref{fig:transformer}, the computations within a layer of most decoder-only LLMs proceed as follows: 
\begin{align}
    f^l &= \texttt{LN1}(h^{l-1}) \label{eq:ln1} \\
    a^l &= \texttt{Att}(f^l)\\
    g^l &= \texttt{LN2}(h^{l-1} + a^l) \label{eq:ln2}\\
    m^l &= W^l_{proj} \sigma(W^l_{fc}g^{l}  + b^l_{fc}) + b_{proj}\\
    h^l &= h^{l-1} + a^l + m^l\label{eq:resiudal-recursive}
\end{align}

The intermediate hidden vectors between each layer, $h^l$, are also sometimes called the \textit{residual stream}. Let \texttt{LN1} represent the first LayerNorm function that acts just before the attention module and \texttt{LN2} represent the second LayerNorm just before the MLP module. We abstract out the computations of the attention and MLP modules to focus on the layer normalization blocks.

As the vectors computed in the attention and MLP modules get added to the residual stream at each layer, the residual stream represents a summation of an increasing number of vectors. A non-recursive formula for the residual stream depicts this clearly:

\begin{equation}\label{eq:resiudal-summation}
    h^l = h^{0} + \sum^{i = l}_{i = 0} a^i + \sum^{i = l}_{i = 0} m^i
\end{equation}

The residual stream thus represents a continuous summation of vectors computed at the attention and MLP modules in each layer with the incoming residual stream. This leads to an increasing norm of the residual stream (section \ref{sec:norm-stabilization}), thus necessitating a normalization operation.

\subsection{Hyperparameters and Computation Resources}

This appendix includes detailed tables and figures that illustrate the effects of layer normalization on the hidden and post-layer normalization vectors across various models.

All experiments were conducted using NVIDIA A6000 GPUs with 48GB of GPU memory. This setup provided the necessary computational power to handle large-scale models and extensive data processing required for our study.

\begin{figure*}[t]
    \centering
    \begin{subfigure}[b]{0.24\textwidth}
        \centering
        \includegraphics[width=\textwidth]{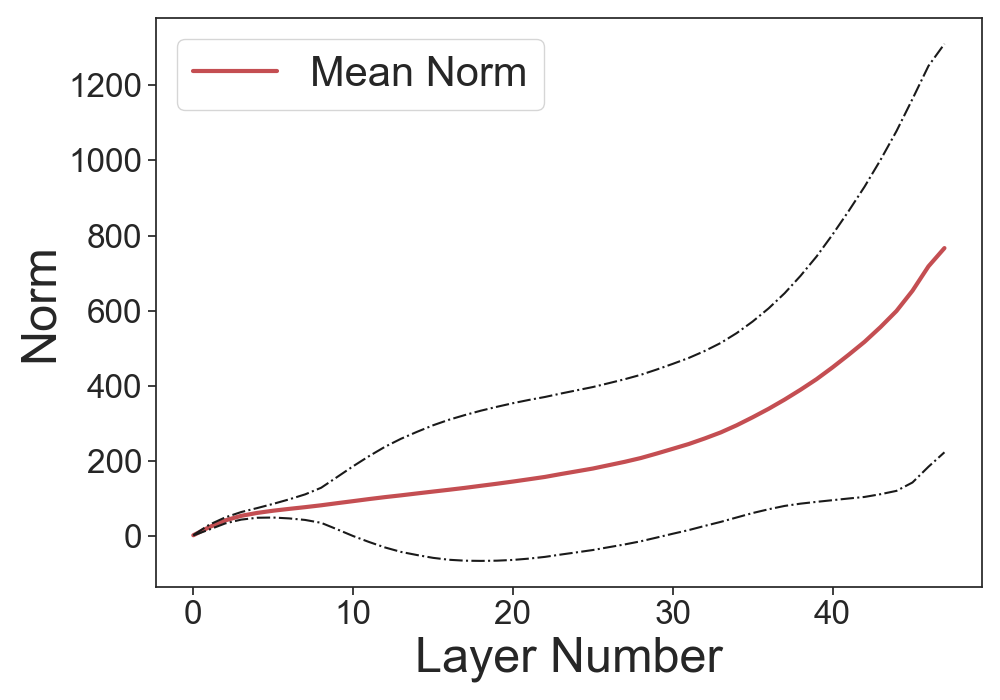}
        \caption{Residual Stream Pre-LN1 GPT-2 XL}
        \label{fig:residual-norm-prelnGPT2xl}
    \end{subfigure}
    \begin{subfigure}[b]{0.24\textwidth}
        \centering
        \includegraphics[width=\textwidth]{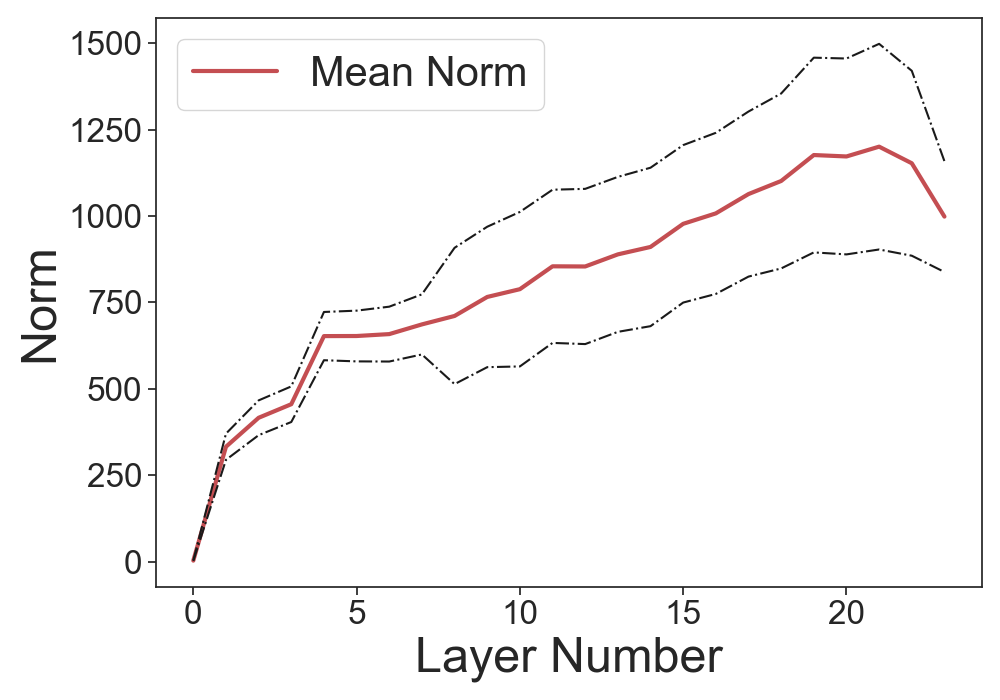}
        \caption{Residual Stream Pre-LN1 GPT-Neo}
        \label{fig:residual-norm-prelnGPTneo}
    \end{subfigure}
    \begin{subfigure}[b]{0.24\textwidth}
        \centering
        \includegraphics[width=\textwidth]{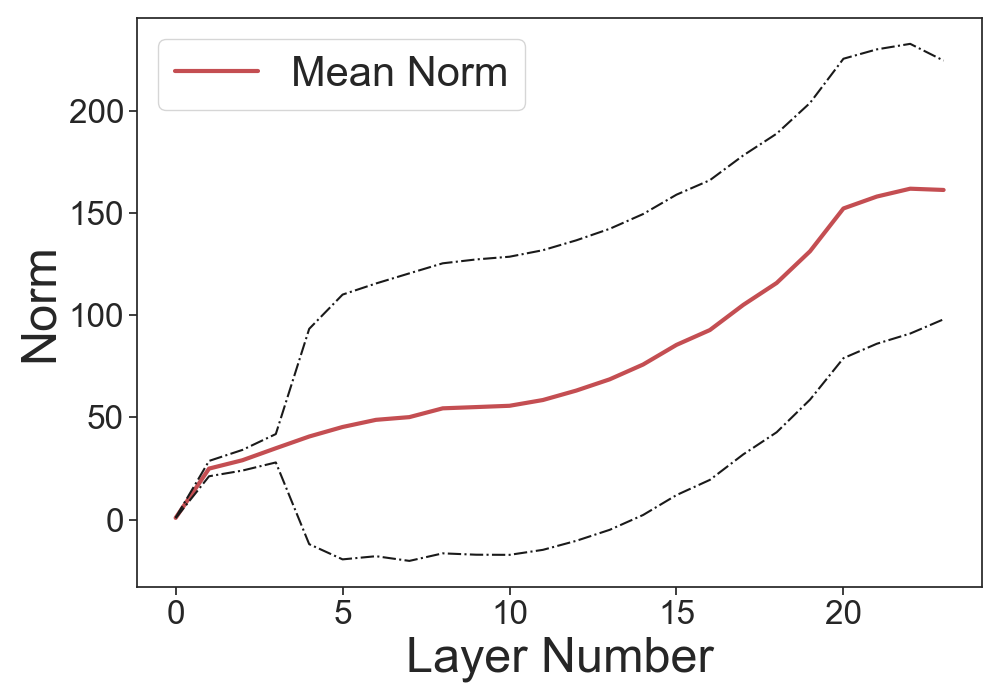}
        \caption{Residual Stream Pre-LN1 Pythia 1.4}
        \label{fig:residual-norm-preln1.4}
    \end{subfigure}
    \begin{subfigure}[b]{0.24\textwidth}
        \centering
        \includegraphics[width=\textwidth]{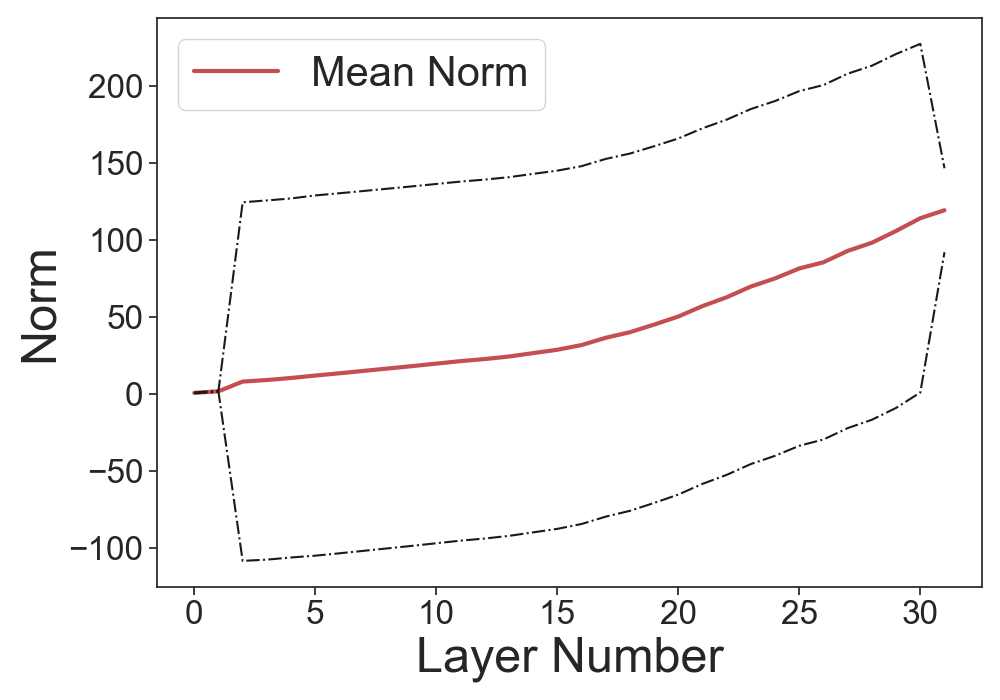}
        \caption{Residual Stream Pre-LN1 Llama-2}
        \label{fig:residual-norm-prelnLLAMA2}
    \end{subfigure}
    \hfill
    \begin{subfigure}[b]{0.24\textwidth}
        \centering
        \includegraphics[width=\textwidth]{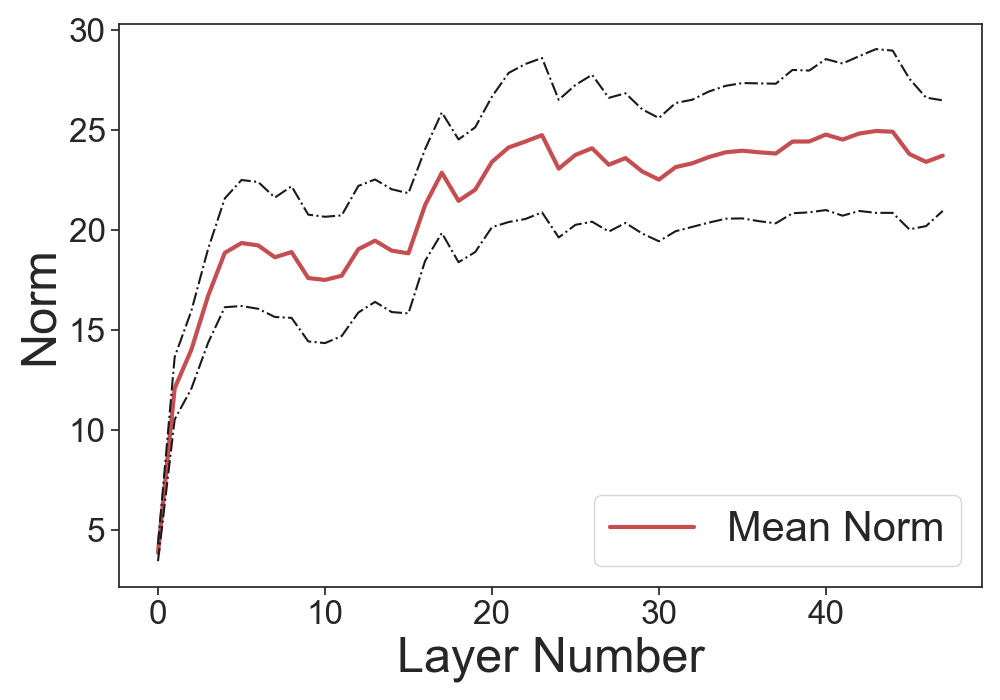}
        \caption{Residual Stream Post-LN1 GPT-2 XL}
        \label{fig:residual-norm-postlngpt2}
    \end{subfigure}
    \begin{subfigure}[b]{0.24\textwidth}
        \centering
        \includegraphics[width=\textwidth]{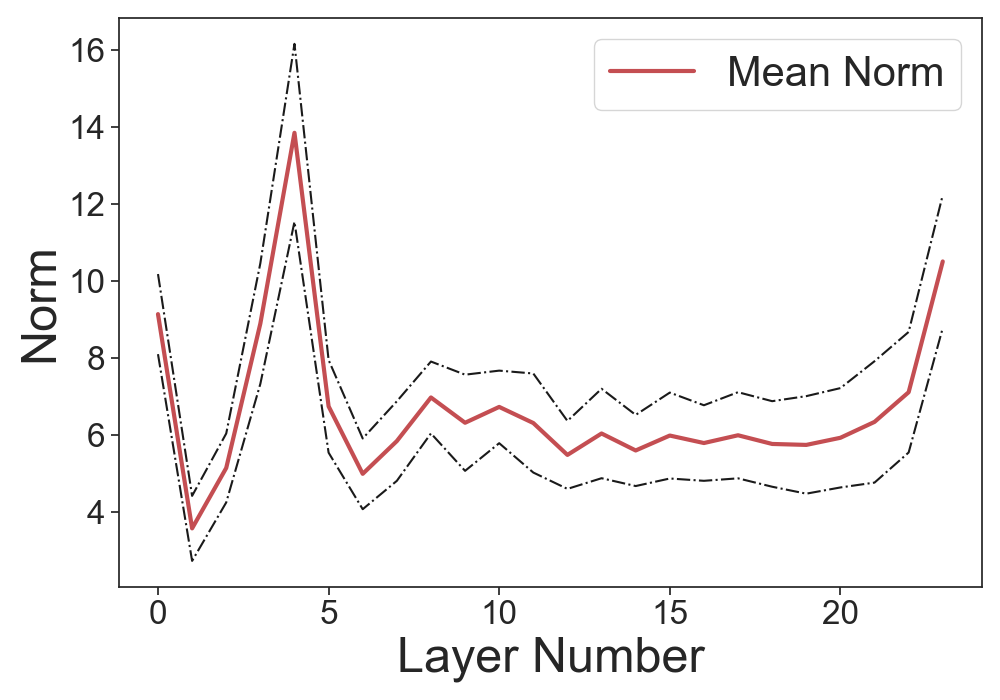}
        \caption{Residual Stream Post-LN1 GPT-Neo}
        \label{fig:residual-norm-postlngptneo}
    \end{subfigure}
    \begin{subfigure}[b]{0.24\textwidth}
        \centering
        \includegraphics[width=\textwidth]{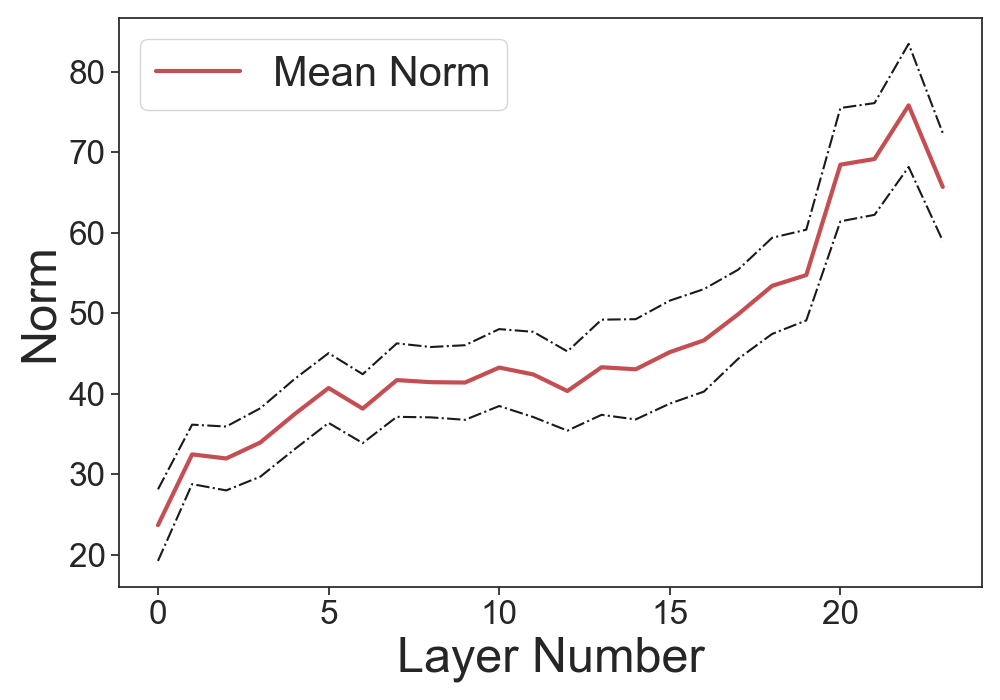}
        \caption{Residual Stream Post-LN1 Pythia 1.4}
        \label{fig:residual-norm-postln1.4}
    \end{subfigure}
    \begin{subfigure}[b]{0.24\textwidth}
        \centering
        \includegraphics[width=\textwidth]{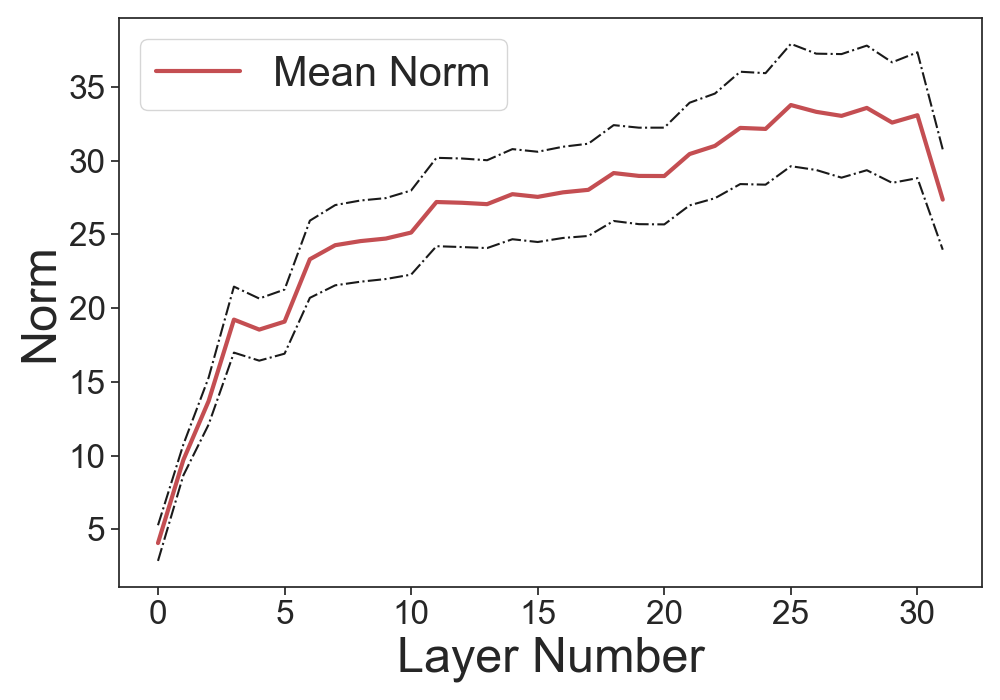}
        \caption{Residual Stream Post-LN1 Llama-2}
        \label{fig:residual-norm-postlnLLAMA2}
    \end{subfigure}
    \caption{Pre and Post-LN1 plots of GPT-2 XL, GPT-Neo, Pythia 1.4 and Llama-2}
    \label{fig:residual-norm-ln-effect_rest}
\end{figure*}

\begin{figure*}[t]
    \centering
    \begin{subfigure}[b]{0.32\textwidth}
        \centering
        \includegraphics[width=\textwidth]{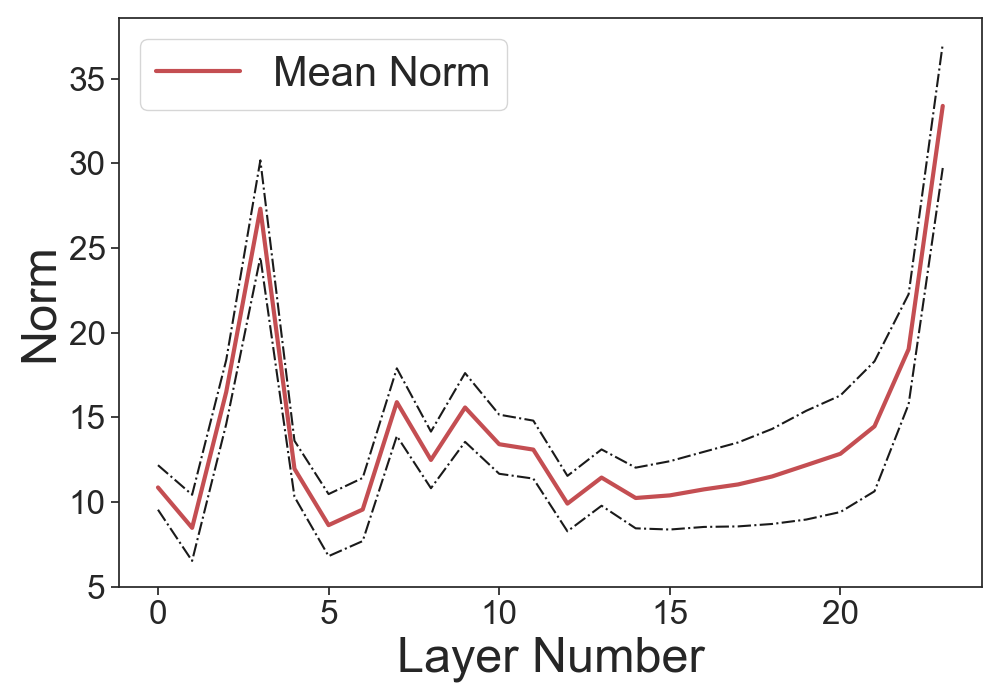}
        \caption{Residual Stream Post-LN2 GPT-Neo}
        \label{fig:residual-norm-postln2GPTneo}
    \end{subfigure}
    \centering
    \begin{subfigure}[b]{0.32\textwidth}
        \centering
        \includegraphics[width=\textwidth]{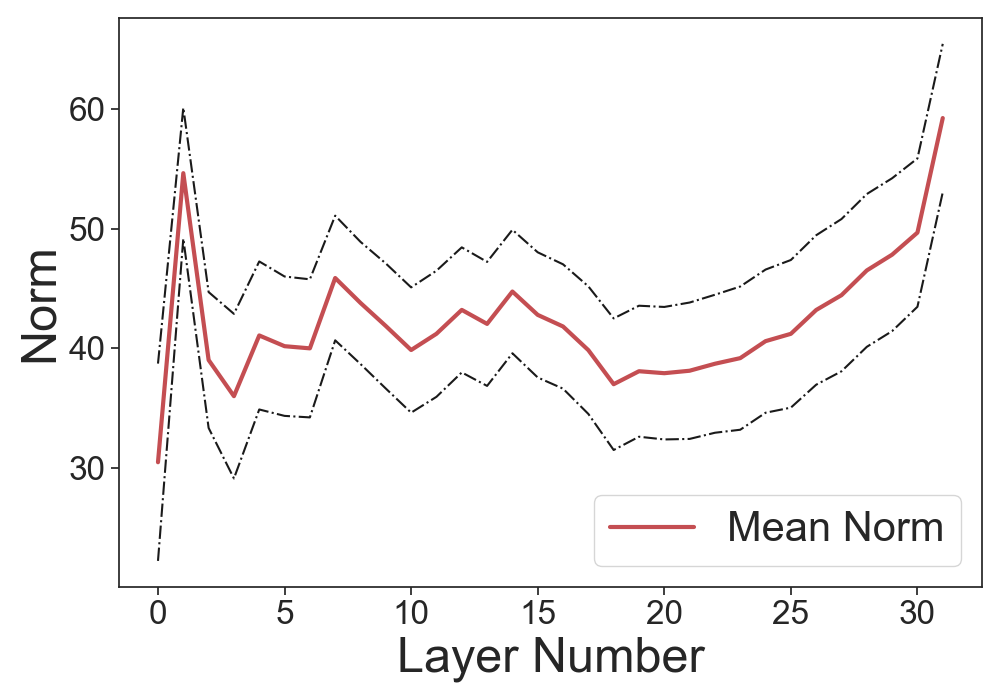}
        \caption{Residual Stream Post-LN2 Pythia 6.9}
        \label{fig:residual-norm-postln26.9}
    \end{subfigure}
    \centering
    \begin{subfigure}[b]{0.32\textwidth}
        \centering
        \includegraphics[width=\textwidth]{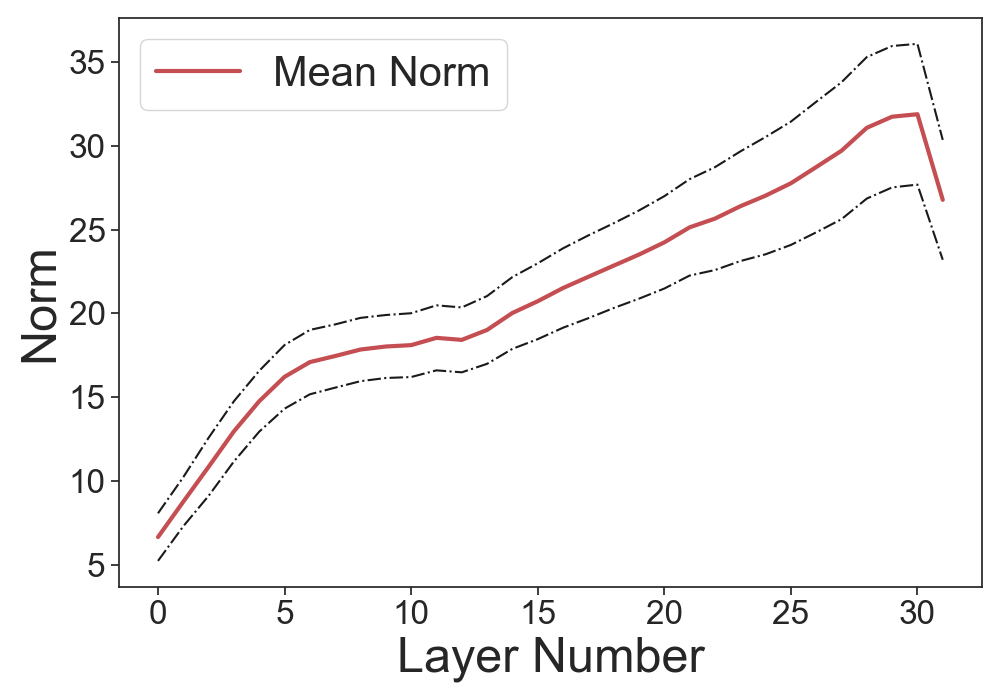}
        \caption{Residual Stream Post-LN2 Llama-3}
        \label{fig:residual-norm-postln2LLAMA3}
    \end{subfigure}
    \hfill
    \begin{subfigure}[b]{0.32\textwidth}
        \centering
        \includegraphics[width=\textwidth]{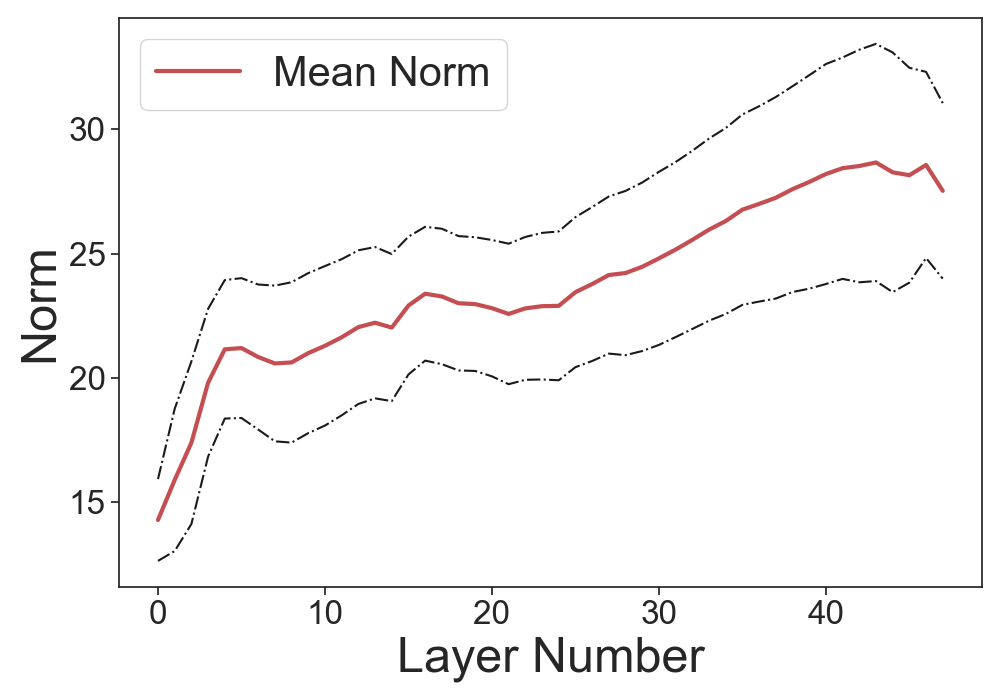}
        \caption{Residual Stream Post-LN2 GPT-2 XL}
        \label{fig:residual-norm-postln2gpt2}
    \end{subfigure}
    \begin{subfigure}[b]{0.32\textwidth}
        \centering
        \includegraphics[width=\textwidth]{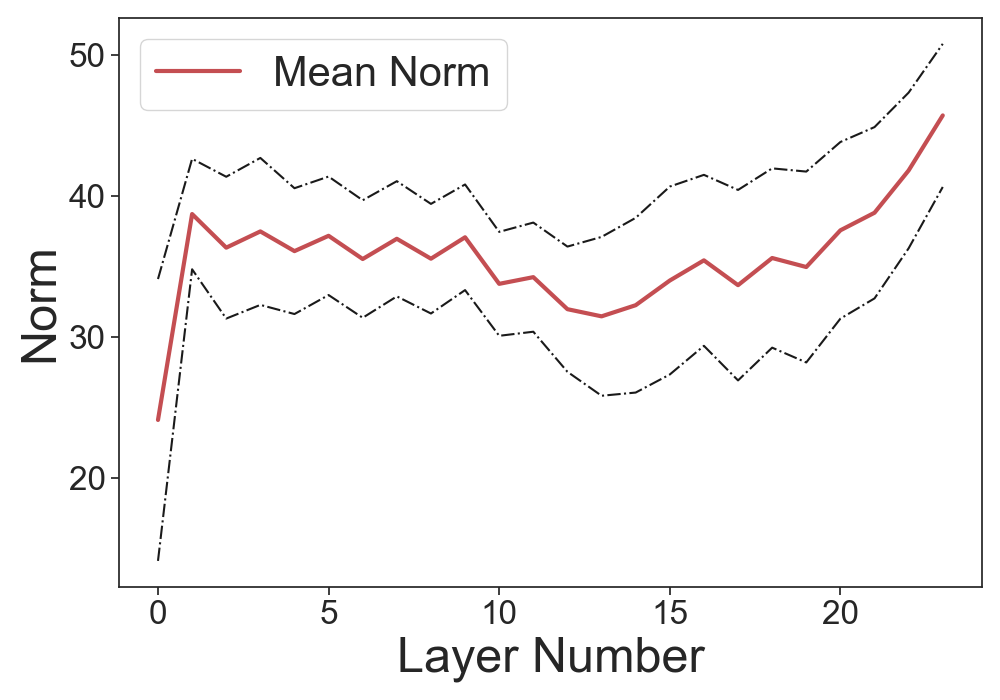}
        \caption{Residual Stream Post-LN2 Pythia 1.4}
        \label{fig:residual-norm-postln21.4}
    \end{subfigure}
    \begin{subfigure}[b]{0.32\textwidth}
        \centering
        \includegraphics[width=\textwidth]{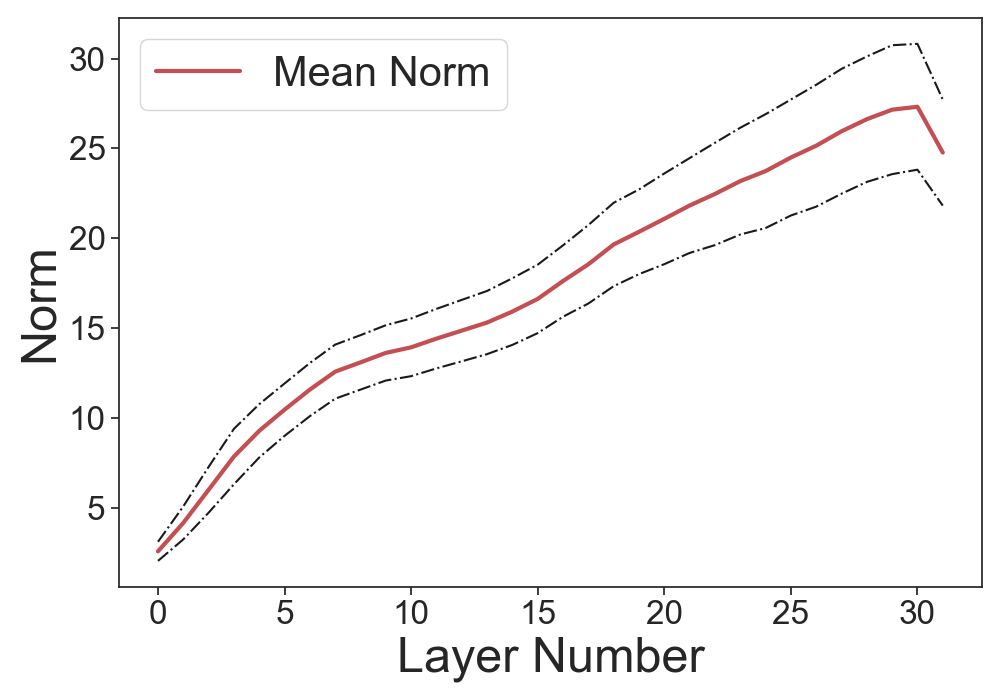}
        \caption{Residual Stream Post-LN2 Llama-2}
        \label{fig:residual-norm-postln2LLAMA2}
    \end{subfigure}
    \caption{Affect of LN2 on stabilizing the norm in all models}
    \label{fig:residual-norm-ln2-effect_all}
\end{figure*}

\begin{figure*}[t]
\centering
    \begin{subfigure}[b]{0.32\textwidth}
        \centering
        \includegraphics[width=\textwidth]{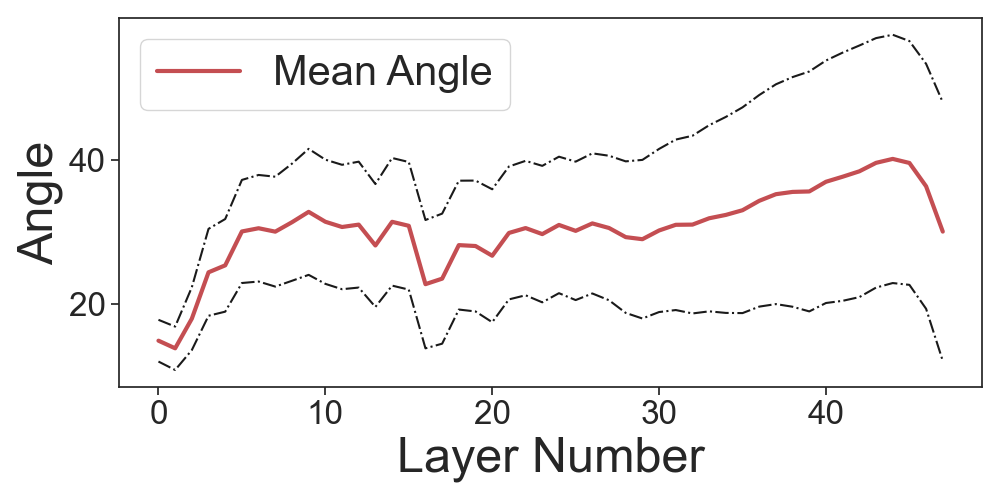}
        \caption{GPT-2 XL}
        \label{fig:rotGPT2XLln1}
    \end{subfigure}
    \centering
    \begin{subfigure}[b]{0.32\textwidth}
        \centering
        \includegraphics[width=\textwidth]{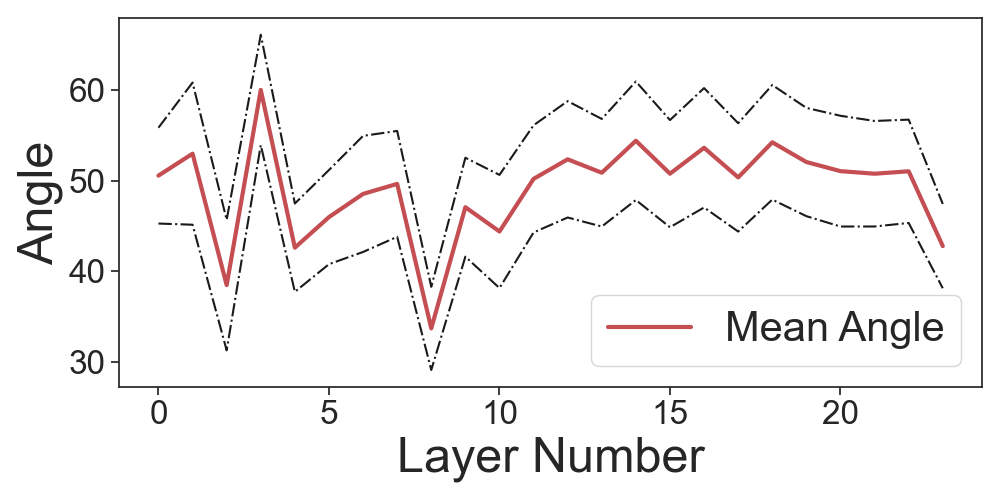}
        \caption{GPT-Neo}
        \label{fig:rotGPTNEOln1}
    \end{subfigure}
    \begin{subfigure}[b]{0.32\textwidth}
        \centering
        \includegraphics[width=\textwidth]{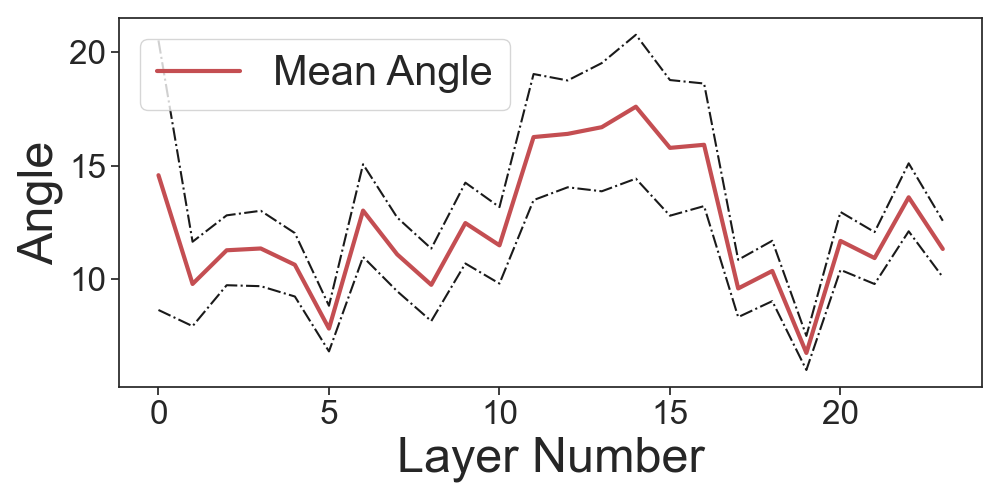}
        \caption{Pythia 1.4}
        \label{fig:rot1.4ln1}
    \end{subfigure}
    \begin{subfigure}[b]{0.32\textwidth}
        \centering
        \includegraphics[width=\textwidth]{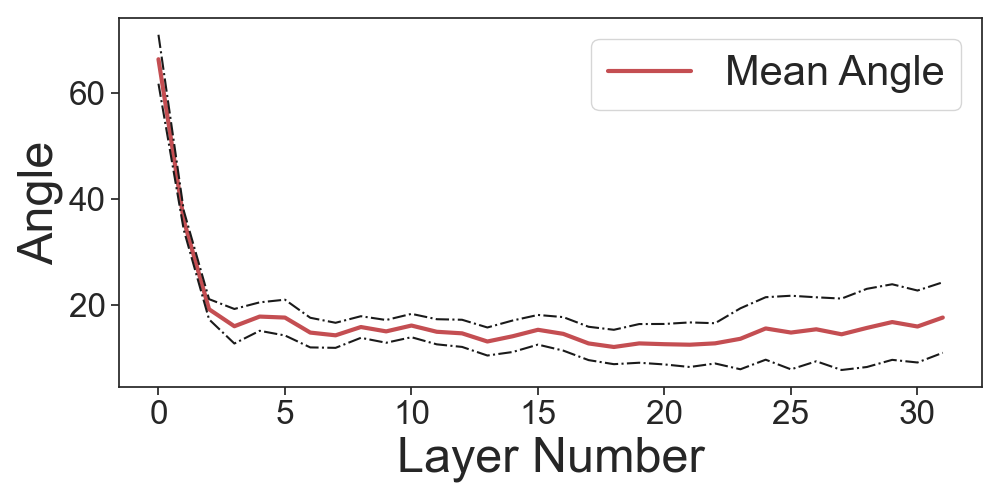}
        \caption{Llama-2}
        \label{fig:rotLLAMA2ln1}
    \end{subfigure}
    \caption{Rotation angle (in degrees) between the hidden vectors and Post-LN1 vectors across all layers for GPT-2 XL, GPT-Neo, Pythia 1.4 and Llama-2}
    \label{fig:rotationanglefigurerest}
\end{figure*}

\begin{figure*}[t]

    \centering
    \begin{subfigure}[b]{0.32\textwidth}
        \centering
        \includegraphics[width=\textwidth]{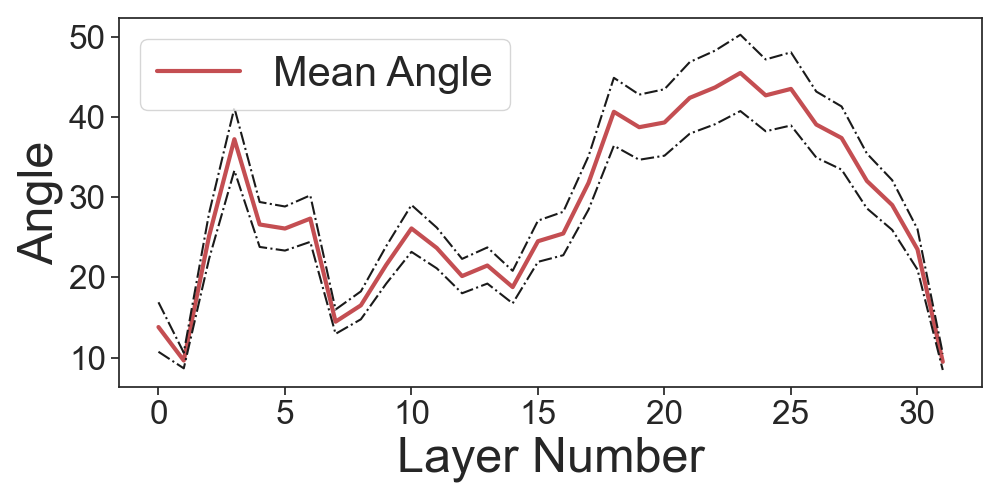}
        \caption{Pythia 6.9}

    \end{subfigure}
    \begin{subfigure}[b]{0.32\textwidth}
        \centering
        \includegraphics[width=\textwidth]{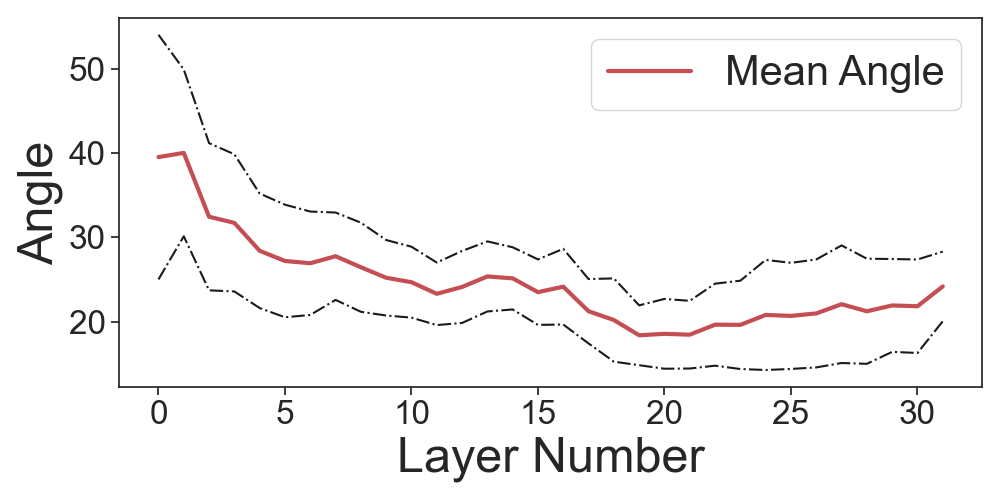}
        \caption{Llama-3}

    \end{subfigure}
    \begin{subfigure}[b]{0.32\textwidth}
        \centering
        \includegraphics[width=\textwidth]{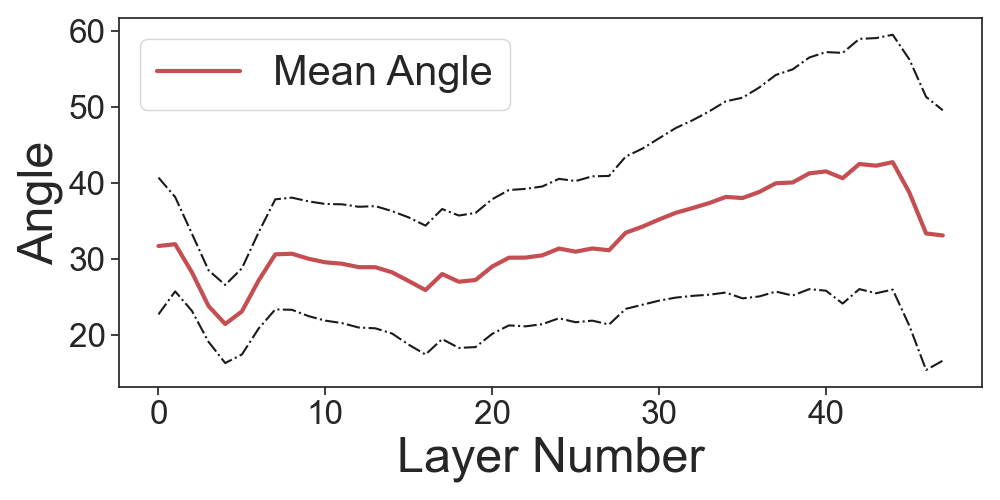}
        \caption{GPT-2 XL}

    \end{subfigure}
    \begin{subfigure}[b]{0.32\textwidth}
        \centering
        \includegraphics[width=\textwidth]{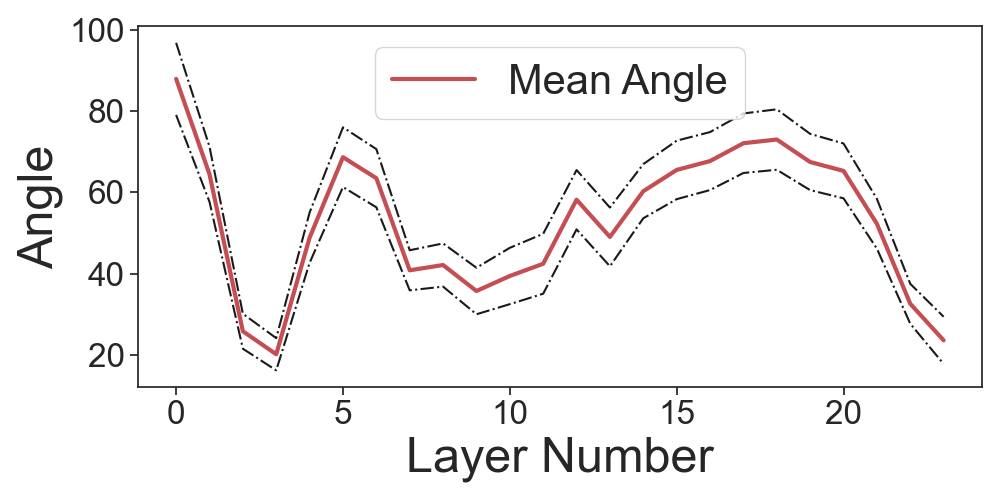}
        \caption{GPT-Neo}

    \end{subfigure}
    \begin{subfigure}[b]{0.32\textwidth}
        \centering
        \includegraphics[width=\textwidth]{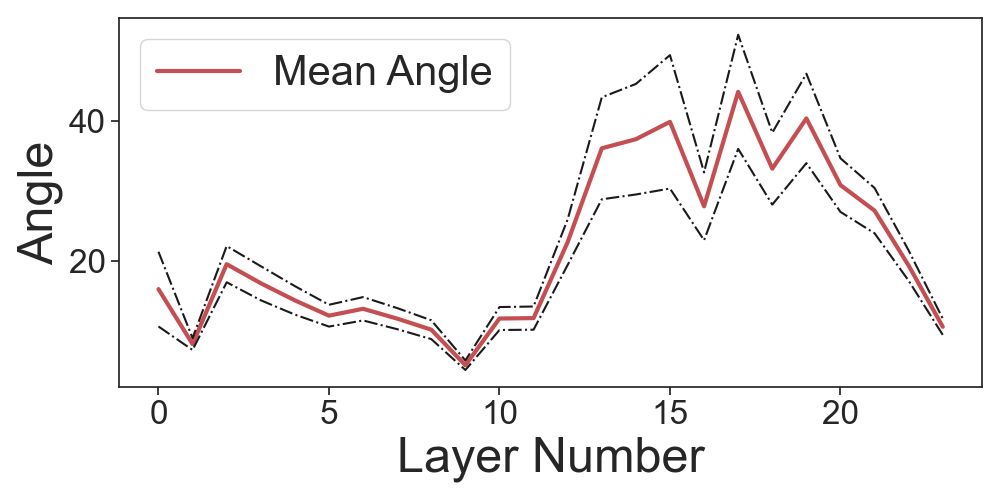}
        \caption{Pythia 1.4}

    \end{subfigure}
    \begin{subfigure}[b]{0.32\textwidth}
        \centering
        \includegraphics[width=\textwidth]{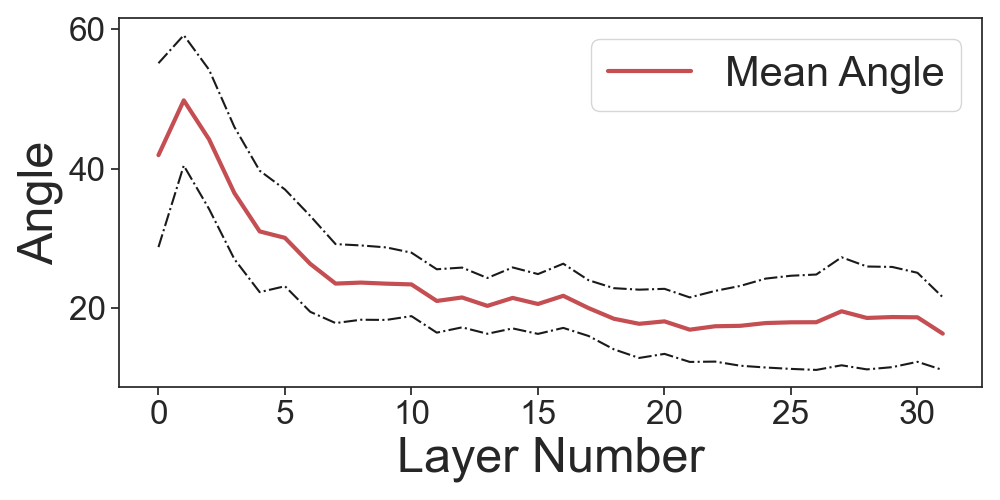}
        \caption{Llama-2}

    \end{subfigure}
    \caption{Rotation angle (in degrees) between the hidden vectors and Post-LN2 vectors across all layers for all models}
    \label{fig:rotationanglefigureallln2}
\end{figure*}

\begin{figure*}[t]
    \centering
    \begin{subfigure}[b]{0.24\textwidth}
        \centering
        \includegraphics[width=\textwidth]{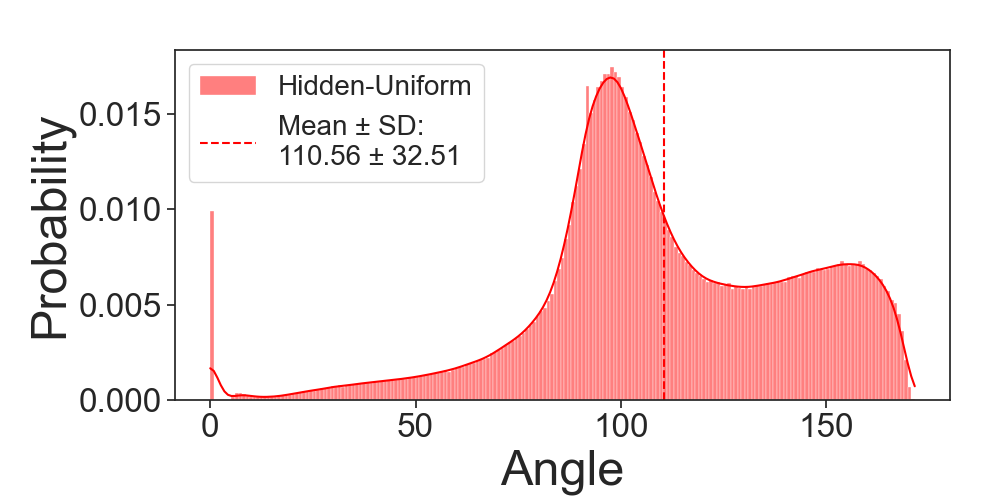}
        \caption{Hidden GPT-2 XL}
    \end{subfigure}
    \begin{subfigure}[b]{0.24\textwidth}
        \centering
        \includegraphics[width=\textwidth]{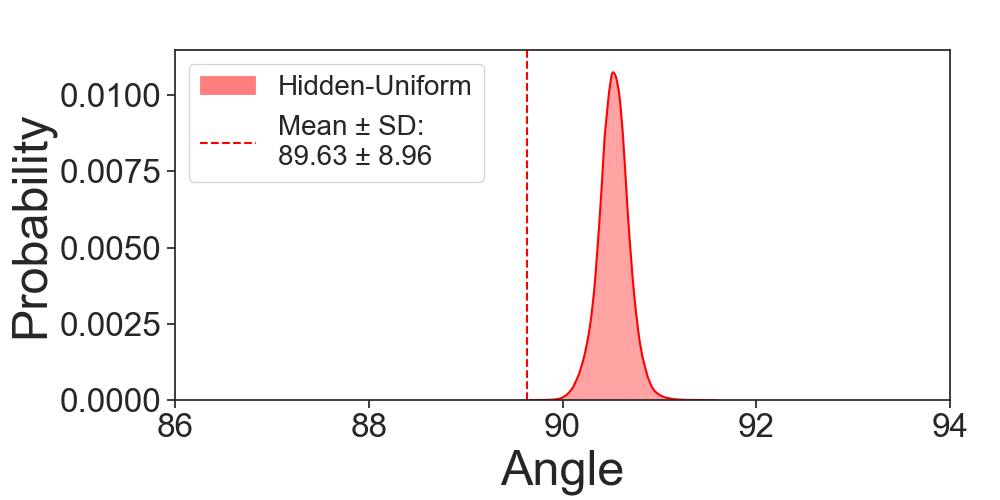}
        \caption{Hidden GPT-Neo}
    \end{subfigure}
    \begin{subfigure}[b]{0.24\textwidth}
        \centering
        \includegraphics[width=\textwidth]{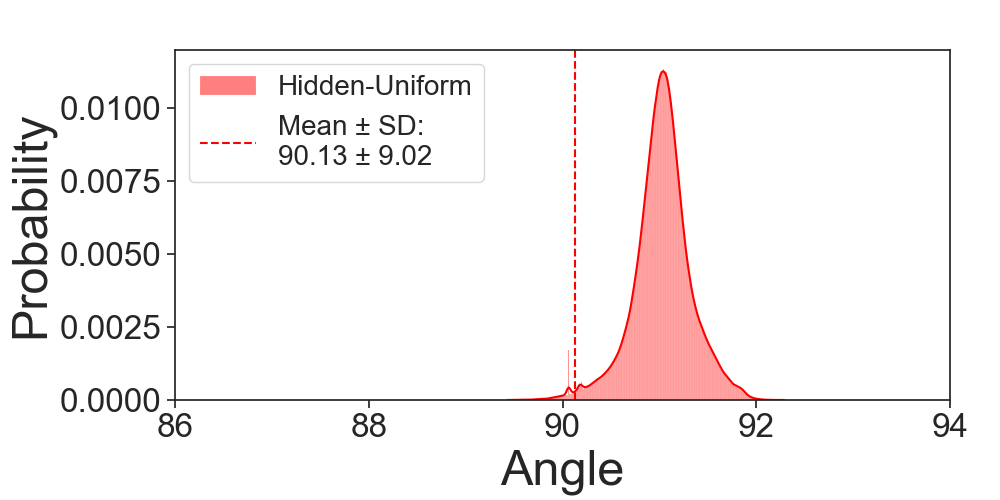}
        \caption{Hidden Pythia 1.4}
    \end{subfigure}
    \begin{subfigure}[b]{0.24\textwidth}
        \centering
        \includegraphics[width=\textwidth]{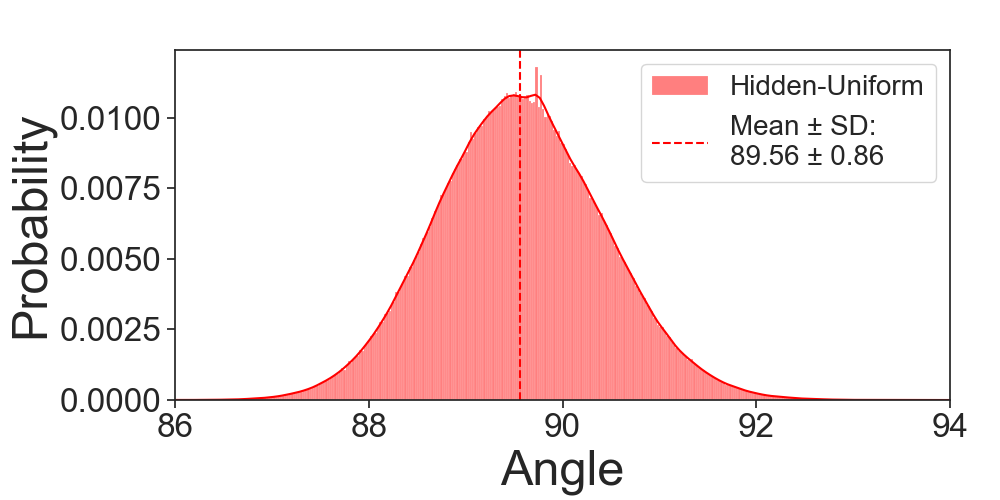}
        \caption{Hidden Llama-2}
    \end{subfigure}

    \hfill
    \begin{subfigure}[b]{0.24\textwidth}
        \centering
        \includegraphics[width=\textwidth]{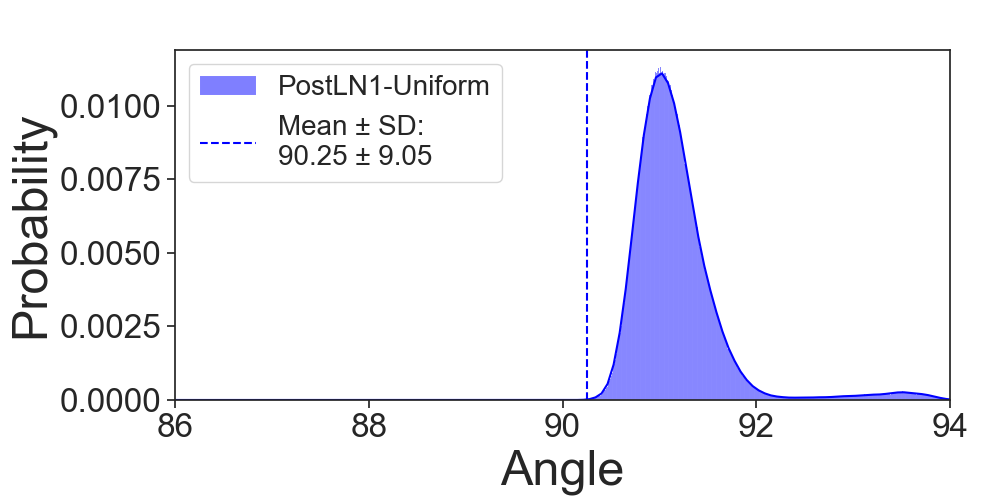}
        \caption{Post-LN1 GPT-2 XL}
    \end{subfigure}
    \begin{subfigure}[b]{0.24\textwidth}
        \centering
        \includegraphics[width=\textwidth]{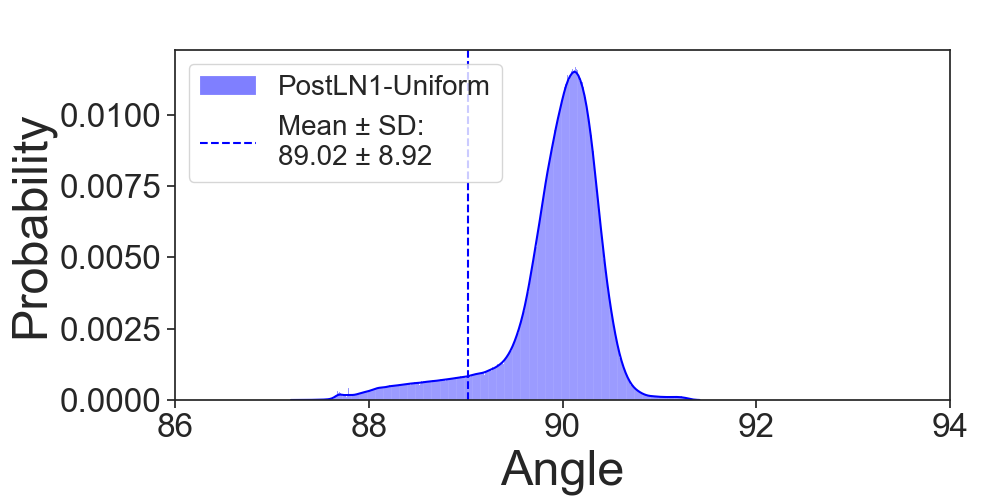}
        \caption{Post-LN1 GPT-Neo}
    \end{subfigure}
    \begin{subfigure}[b]{0.24\textwidth}
        \centering
        \includegraphics[width=\textwidth]{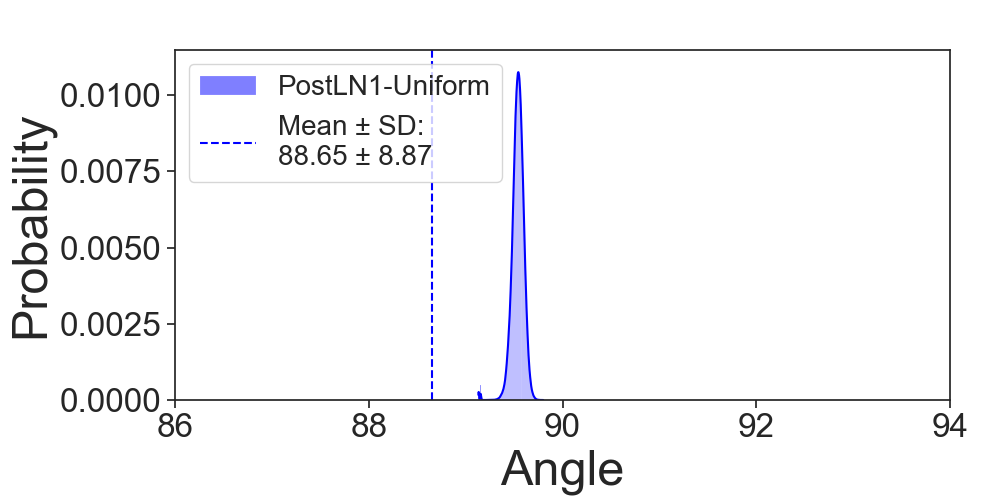}
        \caption{Post-LN1 Pythia 1.4}
    \end{subfigure}
    \begin{subfigure}[b]{0.24\textwidth}
        \centering
        \includegraphics[width=\textwidth]{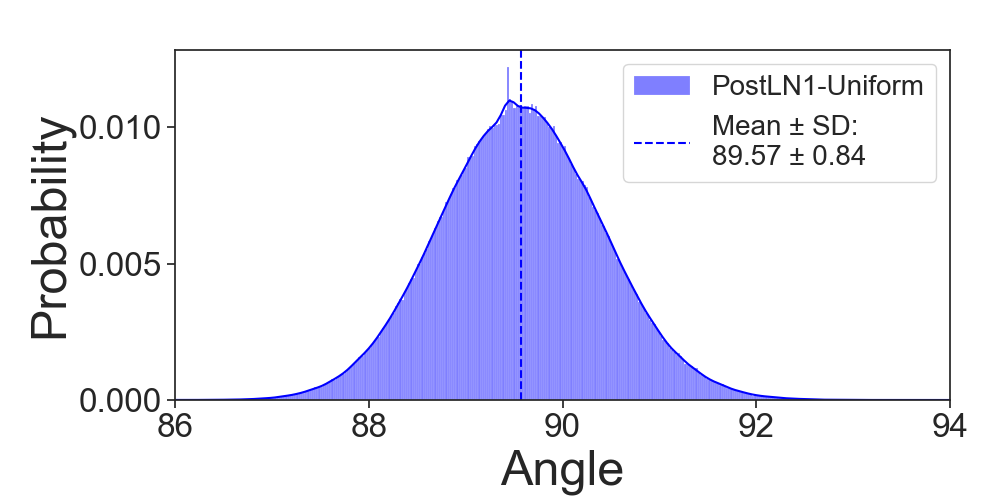}
        \caption{Post-LN1 Llama-2}
    \end{subfigure}

    \caption{Distribution of angles (in degrees) between Hidden vectors (a-d) and post-normalization vectors (e-h) with
the uniform vector for GPT-2 XL (Layer 43), GPT-Neo (Layer 18), Pythia 1.4 (Layer 18), and Llama-2 (Layer 24) for a specific layer chosen}
    \label{fig:angle-diff-main-paperREST}
\end{figure*}

Plots in Figure \ref{fig:residual-norm-ln-effect_rest} compare the norms of the residual streams before and after LN1 across the rest of the models not included in Figure \ref{fig:residual-norm-ln-effect}, demonstrating the normalization's effect. Plots in Figure \ref{fig:residual-norm-ln2-effect_all}, on the other hand, display the effect of LN2 on the residual stream norms across all models, highlighting the further stabilization achieved. Figure \ref{fig:rotationanglefigurerest} shows the rotation angles between hidden vectors and post-LN1 vectors for the remaining models, illustrating the angular changes due to the first layer normalization. Figure \ref{fig:rotationanglefigureallln2} presents the rotation angles between hidden vectors and post-LN2 vectors for all models, and Figure \ref{fig:angle-diff-main-paperREST} presents the angle frequency plots for the rest of the models.

\subsection{Additional Ablations}

\begin{table*}
  \centering
  \begin{tabular}{lccc}
    \toprule
        Model & Hidden \& Uniform & Post-LN1 \& Uniform & Post-LN2 \& Uniform \\
    Name & (Mean \( \pm \) Std) & (Mean \( \pm \) Std) & (Mean \( \pm \) Std) \\
    \midrule
    GPT-2 XL      & 101.48 \( \pm \) 3.20 & 89.95 \( \pm \) 1.30 & 89.63 \( \pm \) 1.30 \\
    GPT-Neo 1.3B  & 89.63 \( \pm \) 1.84 & 88.85 \( \pm \) 1.82 & 89.47 \( \pm \) 1.84 \\
    Pythia-1.4B   & 89.88 \( \pm \) 1.84 & 88.74 \( \pm \) 1.81 & 89.39 \( \pm \) 1.83\\
    GPT-J 6B      & 88.74 \( \pm \) 1.68 &  89.59 \( \pm \) 1.69  & - \\
    Pythia-6.9B   & 89.27 \( \pm \) 1.58 & 89.02 \( \pm \) 1.57 & 88.97 \( \pm \) 1.57 \\
    Llama-2-7B-hf & 89.47 \( \pm \) 0.15 &89.48 \( \pm \) 0.14  & 89.72 \( \pm \) 0.15\\
    Llama-3-8B    & 90.01 \( \pm \) 0.15 & 90.15 \( \pm \) 0.15 & 89.97 \( \pm \) 0.15 \\
    \bottomrule
  \end{tabular}
  \caption{Average angles in degrees between a uniform vector and vectors from Hidden, Post-LN1, and Post-LN2 layers across all layers for each model}
  \label{tab:angle_with_uniform_table}
\end{table*}

\begin{table*}
  \centering
  \begin{tabular}{lcccccc}
    \toprule
    Model & Hidden & Max & Post-LN1 & Max & Post-LN2 & Max \\
    Name & (Mean \( \pm \) Std) & Hidden & (Mean \( \pm \) Std) &  Post-LN1 & (Mean \( \pm \) Std) & Post-LN2 \\
    \midrule
    GPT-2 XL      & 236.85 \( \pm \) 37.02 & 7186.44 & 21.42 \( \pm \) 0.47 & 37.42 & 23.79 \( \pm \) 0.49 & 38.51 \\
    GPT-Neo 1.3B  & 811.84 \( \pm \) 40.29 & 5961.95 & 6.69 \( \pm \) 0.26 & 32.99 & 13.79 \( \pm \) 0.49 & 50.49 \\
    Pythia-1.4B   & 77.94 \( \pm \) 13.17 & 1373.85 & 46.02 \( \pm \) 1.11  & 81.54 & 35.68 \( \pm \) 1.13 & 56.53 \\
    GPT-J 6B      & 110.88 \( \pm \) 30.02 & 4757.86 & 49.44 \( \pm \) 1.31 & 70.93 & - & - \\
    Pythia-6.9B   & 216.04 \( \pm \) 24.61 & 3119.49 & 79.65 \( \pm \) 1.56 & 133.55 & 42.18 \( \pm \) 1.03 & 65.39 \\
    Llama-2-7B-hf & 44.30 \( \pm \) 19.55 & 3287.86 & 26.21 \( \pm \) 0.56 & 38.06 & 17.27 \( \pm \) 0.42 & 30.76 \\
    Llama-3-8B    & 12.12 \( \pm \) 3.15 & 549.19 & 29.46 \( \pm \) 0.59 & 46.21 & 21.30 \( \pm \) 0.49 & 39.67 \\
    \bottomrule
  \end{tabular}
  \caption{This table presents the average norms (L2) and standard deviations for hidden input vectors and post-layer normalization output vectors, alongside the maximum observed values for various models. "Hidden" is an abbreviation for hidden input vectors.}
  \label{tab:norm_table}
\end{table*}

\begin{table*}
  \centering
  \begin{tabular}{lcc}
    \toprule
    Model & Hidden \& Post-LN1 & Hidden \& Post-LN2 \\
    Name & (Mean \( \pm \) Std) & (Mean \( \pm \) Std) \\
    \midrule
    GPT-2 XL      & 30.46 \( \pm \) 1.68 & 32.68 \( \pm \) 1.60 \\
    GPT-Neo 1.3B  & 49.09 \( \pm \) 1.22 & 52.82 \( \pm \) 1.34  \\
    Pythia-1.4B   & 12.35 \( \pm \) 0.46 & 21.70 \( \pm \) 0.90 \\
    GPT-J 6B      & 19.55 \( \pm \) 0.98 & - \\
    Pythia-6.9B   & 11.02 \( \pm \) 0.26 & 28.66 \( \pm \) 0.58  \\
    Llama-2-7B-hf & 17.13 \( \pm \) 0.77 & 23.48 \( \pm \) 1.16 \\
    Llama-3-8B    & 19.64 \( \pm \) 0.62 & 24.55 \( \pm \) 1.08 \\
    \bottomrule
  \end{tabular}
  \caption{This table shows the average rotation angles in degrees between hidden and post-layer normalization vectors (LN1 and LN2) across all layers for various models, indicating the typical angular deviation introduced by normalization processes.}
  \label{tab:rotation-table}
\end{table*}

\clearpage

\begin{figure*}[t]
    \centering
    \begin{subfigure}[b]{0.31\textwidth}
        \centering
        \includegraphics[width=\textwidth]{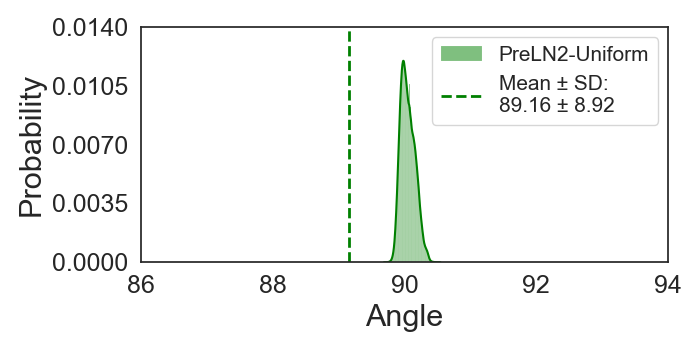}
        \caption{Pre-LN2 Pythia 6.9}
    \end{subfigure}
    \begin{subfigure}[b]{0.31\textwidth}
        \centering
        \includegraphics[width=\textwidth]{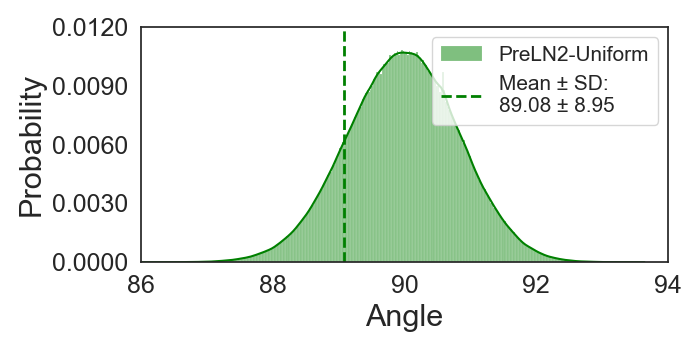}
        \caption{Pre-LN2 Llama-3}
    \end{subfigure}
    \begin{subfigure}[b]{0.31\textwidth}
        \centering
        \includegraphics[width=\textwidth]{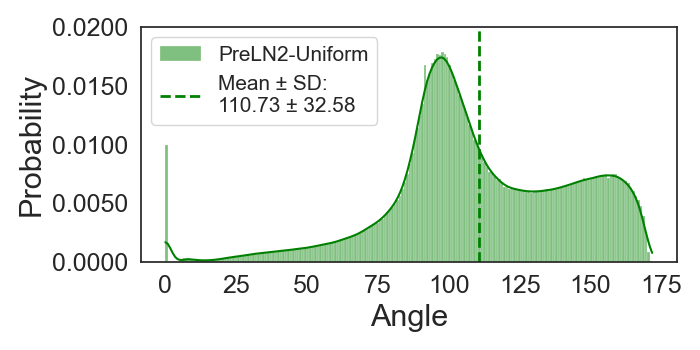}
        \caption{Pre-LN2 GPT-2 XL}
    \end{subfigure}

    \begin{subfigure}[b]{0.31\textwidth}
        \centering
        \includegraphics[width=\textwidth]{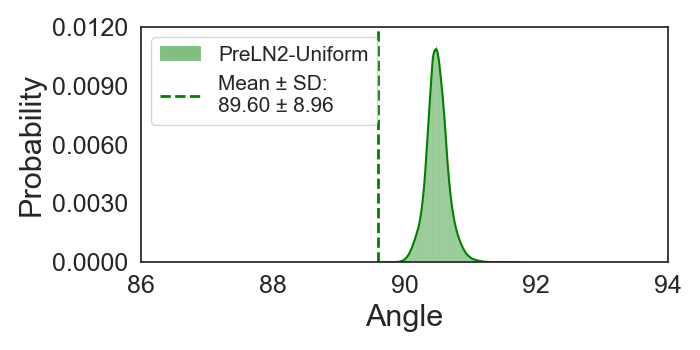}
        \caption{Pre-LN2 GPT-Neo}
    \end{subfigure}
    \begin{subfigure}[b]{0.31\textwidth}
        \centering
        \includegraphics[width=\textwidth]{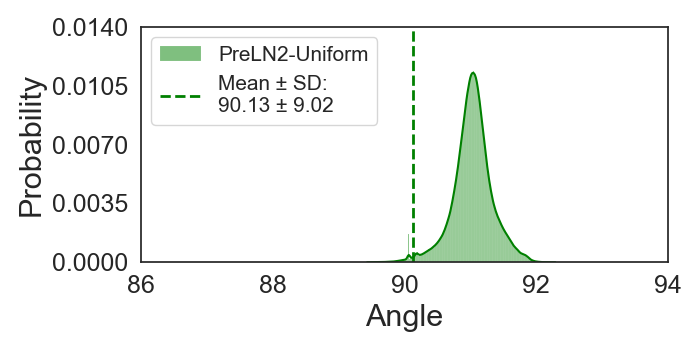}
        \caption{Pre-LN2 Pythia 1.4}
    \end{subfigure}
    \begin{subfigure}[b]{0.31\textwidth}
        \centering
        \includegraphics[width=\textwidth]{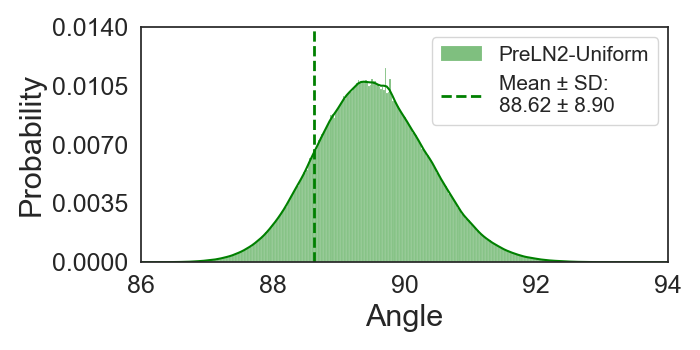}
        \caption{Pre-LN2 Llama-2}
    \end{subfigure}
    \caption{Distribution of angles (in degrees) between (PreLN2 and Uniform) vectors for all models for a single layer chosen (Layer 43 for GPT-2 XL, Layer 18 for GPT-Neo and Pythia 1.4, and Layer 24 for Llama-2, Pythia 6.9, and Llama-3)}
    \label{fig:PRELN2}
\end{figure*}

\begin{figure*}[t]
    \centering
    \begin{subfigure}[b]{0.31\textwidth}
        \centering
        \includegraphics[width=\textwidth]{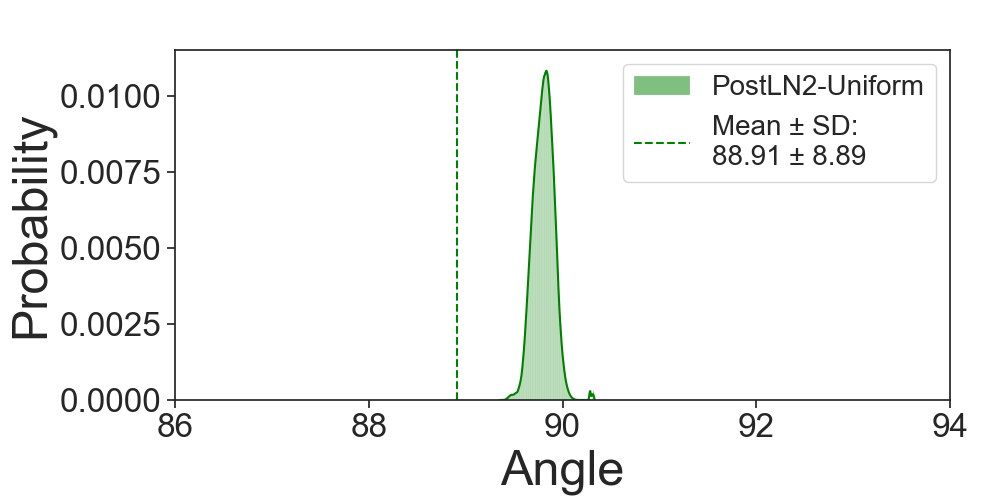}
        \caption{Post-LN2 Pythia 6.9}
    \end{subfigure}
    \begin{subfigure}[b]{0.31\textwidth}
        \centering
        \includegraphics[width=\textwidth]{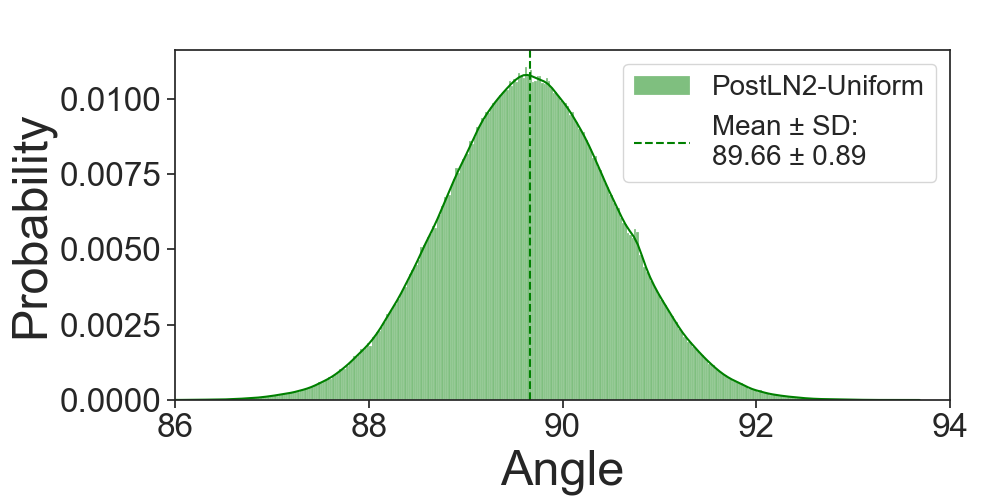}
        \caption{Post-LN2 Llama-3}
    \end{subfigure}
    \begin{subfigure}[b]{0.31\textwidth}
        \centering
        \includegraphics[width=\textwidth]{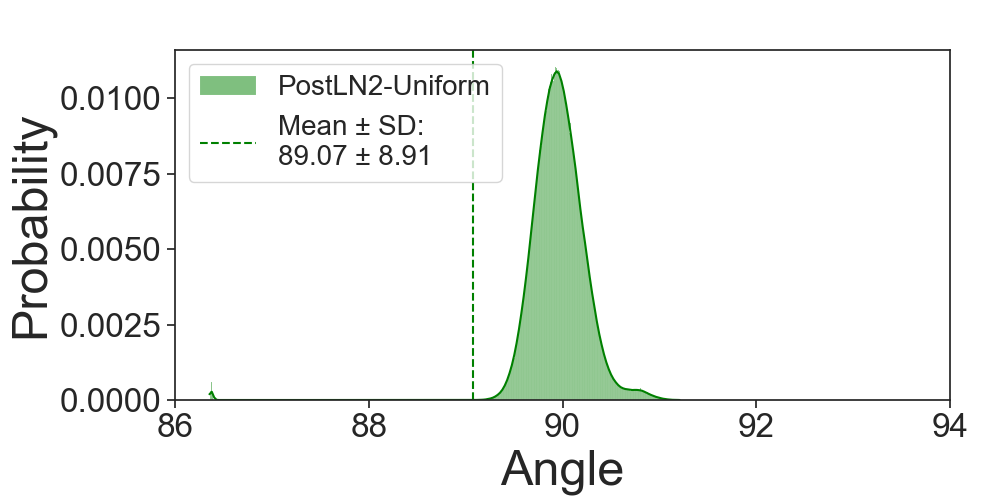}
        \caption{Post-LN2 GPT-2 XL}
    \end{subfigure}

    \begin{subfigure}[b]{0.31\textwidth}
        \centering
        \includegraphics[width=\textwidth]{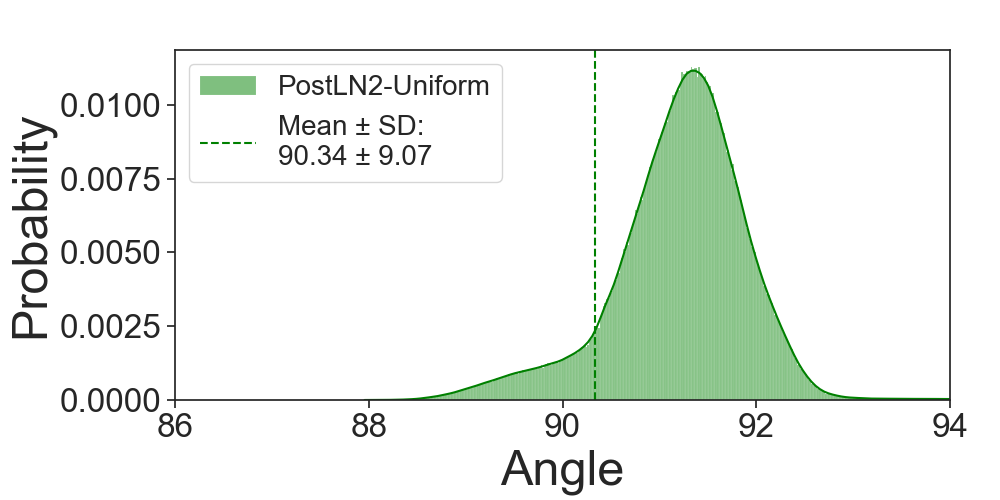}
        \caption{Post-LN2 GPT-Neo}
    \end{subfigure}
    \begin{subfigure}[b]{0.31\textwidth}
        \centering
        \includegraphics[width=\textwidth]{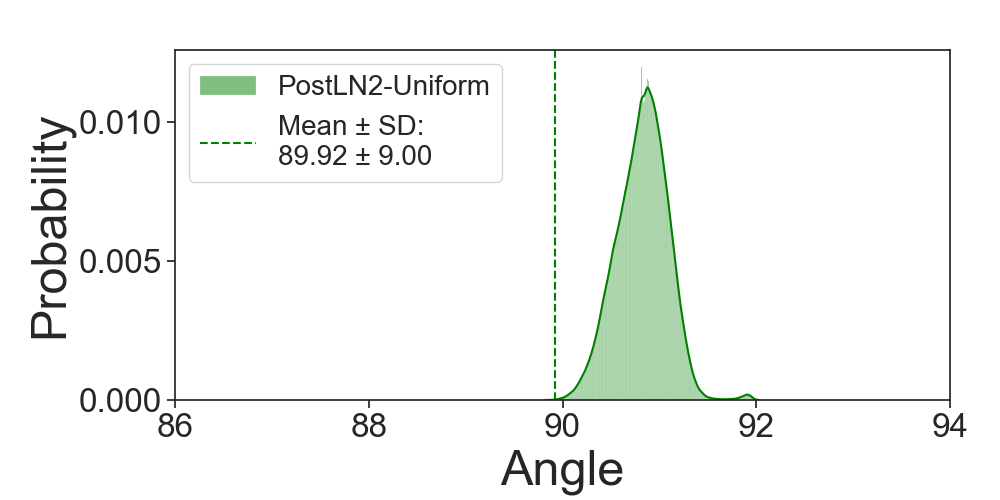}
        \caption{Post-LN2 Pythia 1.4}
    \end{subfigure}
    \begin{subfigure}[b]{0.31\textwidth}
        \centering
        \includegraphics[width=\textwidth]{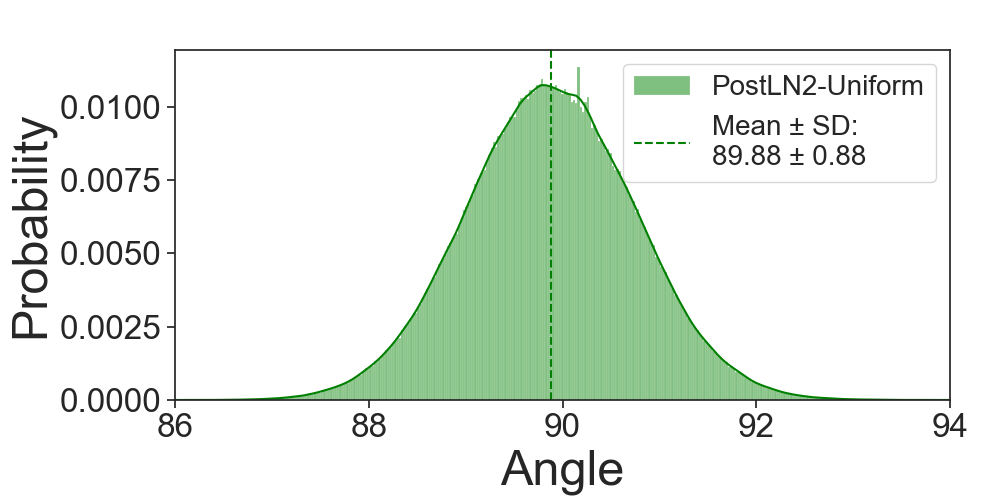}
        \caption{Post-LN2 Llama-2}
    \end{subfigure}
    \caption{Distribution of angles (in degrees) between (PostLN2 and Uniform) vectors for all models for a single layer chosen (Layer 43 for GPT-2 XL, Layer 18 for GPT-Neo and Pythia 1.4, and Layer 24 for Llama-2, Pythia 6.9, and Llama-3)}
    \label{fig:angle-diffLN2-main-paper}
\end{figure*}

\end{document}